\documentclass{article}

\usepackage{microtype}
\usepackage{graphicx}
\usepackage{subcaption}
\usepackage{booktabs} 

\usepackage{hyperref}

\usepackage{amsmath}
\usepackage{amssymb}
\usepackage{mathtools}
\usepackage{amsthm}
\usepackage[dvipsnames]{xcolor}
\usepackage{colortbl}
\usepackage{array}

\usepackage[capitalize,noabbrev]{cleveref}

\usepackage[accepted]{icml2025}

\theoremstyle{plain}
\newtheorem{theorem}{Theorem}[section]

\newtheorem{corollary}[theorem]{Corollary}
\newtheorem{hypothesis}[theorem]{Hypothesis}
\theoremstyle{definition}
\newtheorem{definition}[theorem]{Definition}
\theoremstyle{remark}

\newcommand{\cbt}[2][]{%
  \colorbox[RGB]{255,#2,#2}{\texttt{#1}}%
}
\newcommand{\strongreject}{\textsc{StrongReject} }

\usepackage{natbib}

\usepackage{multirow}
\usepackage{makecell}  
\usepackage{wrapfig}

\usepackage[most]{tcolorbox}
\usepackage{caption}    
\usepackage{tabularx}   
\usepackage{multirow}   
\usepackage{booktabs}   

\icmltitlerunning{The Hidden Dimensions of LLM Alignment}

\begin{document}

\twocolumn[
\icmltitle{The Hidden Dimensions of LLM Alignment: \\ A Multi-Dimensional Analysis of Orthogonal Safety Directions}


\begin{icmlauthorlist}
\icmlauthor{Wenbo Pan}{cityu}
\icmlauthor{Zhichao Liu}{hit}
\icmlauthor{Qiguang Chen}{scir}
\icmlauthor{Xiangyang Zhou}{microsoft}
\icmlauthor{Haining Yu}{hit}
\icmlauthor{Xiaohua Jia}{cityu}
\end{icmlauthorlist}

\icmlaffiliation{cityu}{Department of Computer Science, City University of Hong Kong, Hong Kong}
\icmlaffiliation{hit}{School of Cyberspace Science, Harbin Institute of Technology, China}
\icmlaffiliation{scir}{Research Center for Social Computing and Information Retrieval, Harbin Institute of Technology, China}
\icmlaffiliation{microsoft}{Microsoft}

\icmlcorrespondingauthor{Wenbo Pan}{wenbo.pan@my.cityu.edu.hk}

\icmlkeywords{Machine Learning, Causal Attribution, Jailbreaks, LLMs}

\vskip 0.3in
]

\printAffiliationsAndNotice{}

\begin{abstract}

Large Language Models' safety-aligned behaviors, such as refusing harmful queries, can be represented by linear directions in activation space. Previous research modeled safety behavior with a single direction, limiting mechanistic understanding to an isolated safety feature. In this work, we discover that safety-aligned behavior is jointly controlled by multi-dimensional directions. Namely, we study the vector space of representation shifts during safety fine-tuning on Llama 3 8B for refusing jailbreaks. By studying orthogonal directions in the space, we first find that a dominant direction governs the model's refusal behavior, while multiple smaller directions represent distinct and interpretable features like hypothetical narrative and role-playing. We then measure how different directions promote or suppress the dominant direction, showing the important role of secondary directions in shaping the model's refusal representation. Finally, we demonstrate that removing certain trigger tokens in harmful queries can mitigate these directions to bypass the learned safety capability, providing new insights on understanding safety alignment vulnerability from a multi-dimensional perspective. 
Code and artifacts are available at \url{https://github.com/BMPixel/safety-residual-space}.

\end{abstract}

\section{Introduction}
Large Language Models (LLMs) have demonstrated remarkable capabilities in different domains through extensive pre-training on web-scale text data~\cite{brown2020language, zhao2023survey,qin2024multilingual}. However, toxic content in training data can lead these models to inadvertently generate harmful outputs \cite{su2024mission,gehman2020realtoxicityprompts}. While prior work has aligned LLMs with human preferences~\cite{bai2022training} through safety supervised fine-tuning (SSFT)~\cite{ouyang2022training} and preference optimization like direct preference optimization (DPO)~\cite{rafailov2024direct}, LLMs' safety capabilities can still be bypassed through various attacks, including jailbreak attacks~\cite{zou2023universal,liu2024flipattack,ding2023wolf,yong2023low} and model editing methods \cite{Ball2024UnderstandingJS,arditi2024refusal,carlini2024aligned}. Understanding what models learn during safety fine-tuning is therefore crucial for preventing safety compromises.

\textit{Mechanistic Interpretation}-based methods \cite{bricken2023monosemanticity} have shown promise in explaining safety behaviors of LLMs. These methods study the activation space and identify specific directions that represent meaningful features like toxicity, truthfulness, and refusal \cite{arditi2024refusal, lee2024mechanistic, li2024inference}. However, these directions are typically obtained by training probe vectors on pair-wise datasets (e.g., pairs of safe/unsafe inputs). As a result, the resulting single direction in probe vectors aggregates all contributing signals, potentially conflating different roles of multiple features.

To uncover safety-related directions beyond single-direction probes, we study the activation shift before and after safety fine-tuning, creating a \textit{residual space}. Within this space, we find that safety behavior is controlled by the interplay of multiple safety feature directions. We present a multi-dimensional interpretation of safety mechanisms by explaining each feature direction through its top-contributing training tokens and measuring its effects on other feature directions and safety behaviors. Our contributions are as follows:

\textbf{Introducing Safety Residual Space.} In \autoref{sec:safety_residual}, we define the \emph{safety residual space} as the linear span of representation shifts during safety fine-tuning. We verify that orthogonal directions in this space captures features of alignment goals. In \autoref{sec:linear}, we setup a case study of safety fine-tuning, applying SSFT and DPO on Llama 3 8B for refusing challenging jailbreaks.

\textbf{Discovering Interpretable Directions.} In \autoref{sec:interpretation}, we decompose the space into major directions (i.e., top singular vectors) and extend layer-wise relevance propagation \cite{bach2015pixel} to analyze these directions. We find that a dominant direction governs the model's refusal behavior, while multiple smaller (non-dominant) directions represent distinct and interpretable features such as hypothetical narrative and role-playing. Intervention experiments show that these indirect features regulate different aspects of capabilities learned during safety fine-tuning.

\textbf{Vulnerabilities in Safety Directions.} In \autoref{sec:application}, we examine dynamics in the safety residual space and find the vital role of non-dominant features in promoting dominant direction and refusal. Leverage this insight, we demonstrate that identifying and removing trigger tokens from harmful prompts can reduce refusal even on safety fine-tuned model, thereby circumventing learned safety alignment.

%
\section{Preliminaries}
\label{sec:preliminaries}

\paragraph{Linear Representation} 
We build our framework on the Linear Representation Hypothesis from \citet{park2023linear}. A one-dimensional \emph{feature} value $W$ (e.g., ``gender'', ``harmfulness'') is defined as a latent variable expressed in context $w$. In safety analysis, $w$ typically represents user queries with varying safety aspects - from benign questions like ``What leads to a united society?'' to harmful ones like ``How to make a handgun?''. Probability of feature $W$ presenting in output is denoted as $\mathbb{P}(W)$ (e.g., safe or unsafe responses).

Let $\lambda: w \to \mathbb{R}^d$ be a mapping from context $w$ to its representation.
We say that $\mathbf{x} \in \mathbb{R}^d$ is a \emph{feature direction} of feature $W$ if there exists a pair of contexts $w_0, w_1$ such that $\lambda(w_1) - \lambda(w_0) \in \{\alpha \mathbf{x} : \alpha > 0\}$ satisfying:

\begin{align}
    \frac{\mathbb{P}(W = 1 \mid \lambda(w_1))}{\mathbb{P}(W = 1 \mid \lambda(w_0))} > 1.
    \label{eq:pop}
\end{align}

This inequality ensures that the direction positively contributes to feature $W$.

\paragraph{Safety Directions} 
In LLM safety alignment, researchers have identified distinct feature directions for various safety aspects including bias, toxicity, and refusal behavior. To find such directions, we first construct two sample sets with only difference being $W$ present. The feature direction $\mathbf{v}_W$ is then obtained by maximizing the distance between these two distributions. To verify causality, we can intervene by suppressing this direction in the activation space:

\begin{align}
    \mathbf{x} := \mathbf{x} - \alpha \mathbf{v}_W.
    \label{eq:intervene}
\end{align}

A concrete example is the \emph{refusal direction} identified by \citet{arditi2024refusal}, where they construct contrast pairs by comparing inputs that elicit either compliant or refusing responses.

\paragraph{Layer-wise Relevance Propagation} 
Layer-wise Relevance Propagation (LRP)~\cite{bach2015pixel} decomposes a neural network function $f$ into individual contributions from input variables. For each input-output pair $(i,j)$, we compute a relevance score $R_{i \rightarrow j}$ representing how much input $i$ contributes to output $j$:

\begin{align*}
    f_j(\mathbf{x}) \propto R_j = \sum_{i}^N R_{i \rightarrow j}
\end{align*}

A key property of LRP is conservation across layers. In a layered directed acyclic graph, relevance values $R_i^l \propto f_i$ from a later layer are back-propagated to the previous layer $R_i^{l-1}$ while maintaining constant sum: $\sum_i R_i^{l-1} = \sum_i R_i^l$ In this work, to ensure faithful relevance propagation, we adopt implementation from \citet{achtibat2024attnlrp}.

\begin{figure}
    \vskip 0.2in
    \begin{center}
    \centerline{\includegraphics[width=\columnwidth]{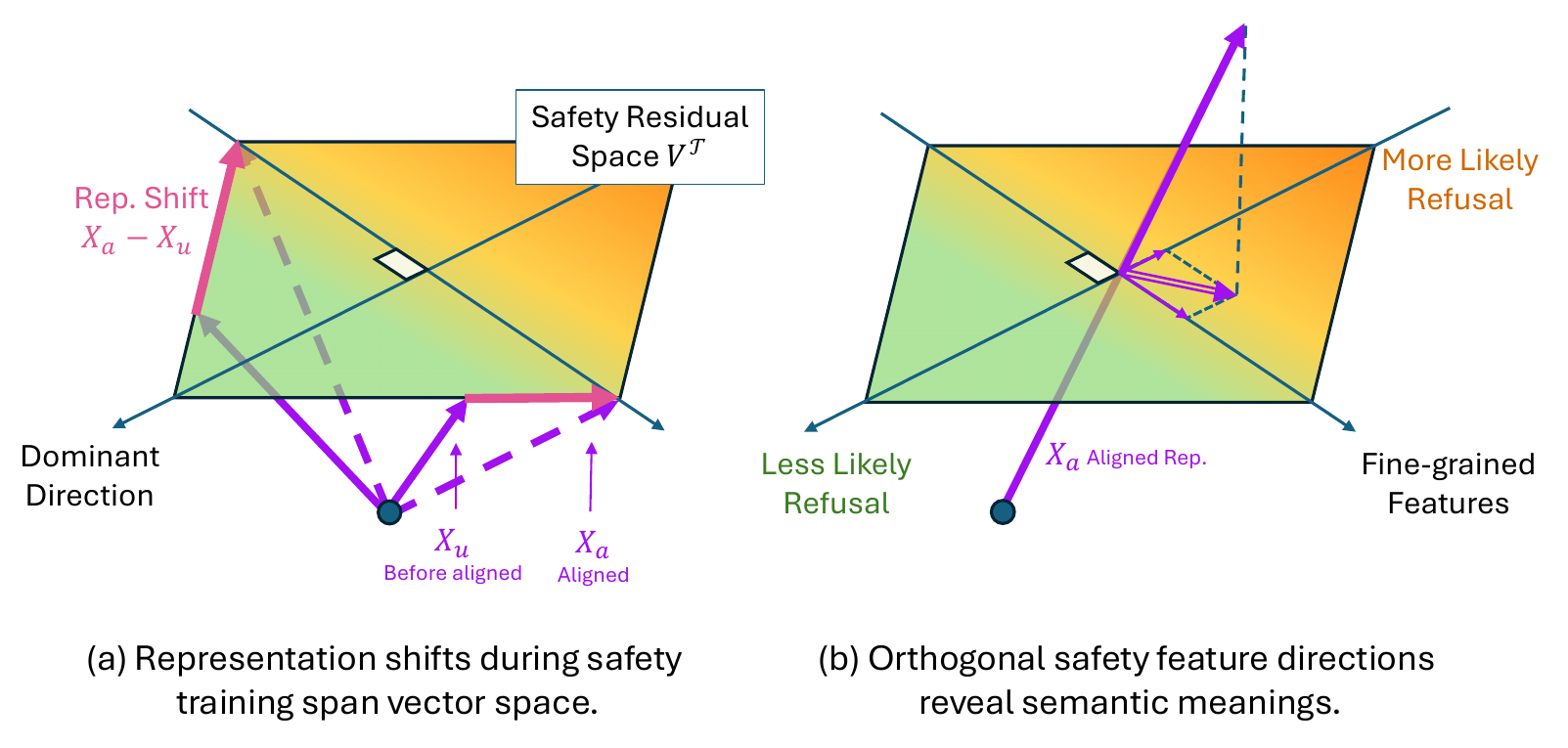}}
    \caption{Illustration of the Safety Residual Space. The safety residual space is the linear span of representation shifts during safety fine-tuning. In our experiments, the dominant direction predicts safety behavior, while non-dominant directions capture different indirect safety features.}
    \label{fig:illustration}
    \end{center}
    \vskip -0.2in
\end{figure}
\section{Safety Residual Space}
\label{sec:safety_residual}

We first define our framework of safety residual space. Our approach is motivated by recent research on training dynamics of alignment algorithms \cite{jain2024makes,lee2024mechanistic}, which shows that representation shifts during training are both meaningful and interpretable. We focus specifically on the effects of safety fine-tuning by comparing representation dynamics before and after the fine-tuning process, limiting our scope to a single forward pass. Let $\mathbf{x}$ denote vectors in the representation space $\mathcal{X} \subset \mathbb{R}^d$. Then $\mathcal{T}: \mathcal{X} \to \mathcal{X}$ describes the representation shift from unaligned (before fine-tuning) to aligned (after fine-tuning) states. We define the safety residual space as the linear span of representation shifts during safety fine-tuning. Formally:

\begin{definition}[Safety Residual Space]
    Consider $\mathcal{T}$ from training on unaligned samples whose representation is $\mathbf{x} \sim \mathbb{P}(\mathcal{X}_u)$.
    The \emph{safety residual space} $\mathcal{S}(\mathbf{x})$ is defined as the optimal affine transformation parameterized as $\mathcal{S}(\mathbf{x}) = \mathbf{W} \mathbf{x} + \mathbf{b}$ that minimizes:
    \[
    \mathcal{S}
    \;=\;
    \underset{\mathbf{\hat{W}},\,\mathbf{\hat{b}}}{\mathrm{argmin}}
    \;\; 
    \mathbb{E}_{\mathbf{x} \sim \mathcal{X}_u}
    \Bigl\|\mathcal{T}(\mathbf{x}) \;-\; \bigl(\mathbf{\hat{W}}\mathbf{x} + \mathbf{\hat{b}}\bigr)\Bigr\|^2.
    \]
    \label{def:residual}
    \end{definition}

Intuitively, this definition captures the linear activation shifts that the model learns from the training. We consider the safety feature directions as linear and ignore non-linear error between $\mathcal{S}$ and $\mathcal{T}$. We use activations from transformer block outputs at the position of the first generated token from each layer. We compute activations from the training data as an approximation of the unaligned distribution $\mathcal{X}_u$. Our experiments show that the $\mathcal{S}$ is a good approximation of the $\mathcal{T}$ with low error, for which we provide the Mean Squared Error (MSE) of $\mathcal{S}$ and $\mathcal{T}$ in \autoref{tab:mse_appendix}.


\paragraph{Extracting Principal Components}
 To identify important directions in the residual space, we apply Singular Value Decomposition (SVD) to $\mathbf{W} - \mathbf{I}$ and take the first $k$ right vectors (components) $V^{:k}$. This describes the span of the largest $k$ orthogonal representation shifts from the input space (i.e., the model before training). 

\paragraph{Notation}
We denote different components as \texttt{LN-CK}, where \texttt{LN} is the layer number and \texttt{K} is the \texttt{K}th largest right vector from SVD. Specifically, we refer to the \texttt{LN-C1} as the \emph{dominant component}, while others are \emph{non-dominant components}. We provide an illustration in \autoref{fig:illustration}. We use \textit{component} and \textit{direction} interchangeably in this paper.

\subsection{Component as Feature Direction}

A key question is whether the components in the residual space contain interpretable features, similar to probe vectors. Conceptually, the safety finetuning optimizes the model to produce safer outputs. This process induces activations to shift along specific directions to align with safety objectives, which we capture with $\mathcal{S}$. These directions in $\mathcal{S}$ are strong candidates for feature directions under the definition in Equation~1, as they increase the probability of safe output when activations are moved along those directions. While this does not guarantee human-interpretable features, it suggests $\mathcal{S}$ is a promising source for automatically discovering safety-related feature directions without requiring probing data pairs. To generalize this idea, we have the following hypothesis:

\begin{hypothesis}[Finetuning Residuals as Feature Directions]
The principal components representing the activation shifts induced by safety finetuning contain safety-related feature directions. Furthermore, orthogonal directions within this space potentially represent distinct and interpretable safety features.
\end{hypothesis}

In the following sections, we verify this hypothesis by examining the top components of $\mathcal{S}$. We study (1) if the components in $\mathcal{S}$ are feature directions and (2) what specific features these directions represent.

\paragraph{Not All Features are in Residuals}

On the other hand, can all features be captured in the residual space? We posit that it primarily reflects features developed during safety training. Features might be absent for two reasons: (1) they are irrelevant to the training objective (e.g., unrelated syntactic patterns), as optimization naturally excludes non-contributing directions; or (2) they already existed in the pre-trained model (e.g., recognizing toxic content), requiring no parameter updates. This implies the residual space spans directions learned during safety fine-tuning.

\begin{corollary}
    The safety residual space is the span of feature directions developed during safety training.
\end{corollary}

\subsection{Experimental Setup}
Now, we describe the experiment setup, focusing on how models learn to recognize and handle unaligned harmful queries through safety fine-tuning.

\paragraph{Dataset} 
We construct a preference dataset of 2600 samples, detailed in \autoref{fig:train_test_set}, incorporating various challenging jailbreak methods and alignment blindspots from recent research \cite{ding2023wolf,yu2023gptfuzzer,zou2023universal,chao2023pair,liu2024flipattack}. This dataset was used both for safety fine-tuning and for learning the safety residual space map. To generate harmful examples, we apply these jailbreak methods to toxic samples from STRONG REJECT~\cite{souly2024strongreject}. We further incorporate 50\% samples from or-bench~\cite{cui2024or} as harmless samples to balance the dataset. All prediction and intervention evaluations were performed on the test set. Detailed dataset specifications are provided in the Appendix~\ref{appd:dataset_construction}.

\paragraph{Evaluation Metrics}
Following established practices in jailbreak research~\cite{zou2023universal,souly2024strongreject}, we evaluate model responses along two dimensions: refusal accuracy and response harmfulness. We measure refusal accuracy across both harmless and harmful test samples, while quantifying response harmfulness using STRONG REJECT scores~\cite{souly2024strongreject}.

\paragraph{Safety Fine-tuning}
We perform safety fine-tuning on Llama 3.1 8B Instruct using both SSFT and DPO approaches for one epoch. For SSFT, we follow \citet{inan2023llama} to optimize the model to generate refusal for harmful queries with instruction fine-tuning. For DPO, we additionally create a preference dataset with prefered helpful responses for harmless queries and refusals with disclaimers for harmful queries. We use Llama 3.1 405B Instruct~\cite{dubey2024llama} to generate reference responses for the preference dataset. The effectiveness of our fine-tuning process is demonstrated by two key metrics: the average \strongreject score~\cite{souly2024strongreject} across all jailbreak attempts decreased significantly from 0.65 to 0.05, while the refusal accuracy improved to 90\%. We provide more details and results on different model sizes in the Appendix~\ref{appd:training_procedure}.
\begin{figure}
    \vskip 0.2in
    \begin{center}
    \center{\includegraphics[width=\columnwidth]{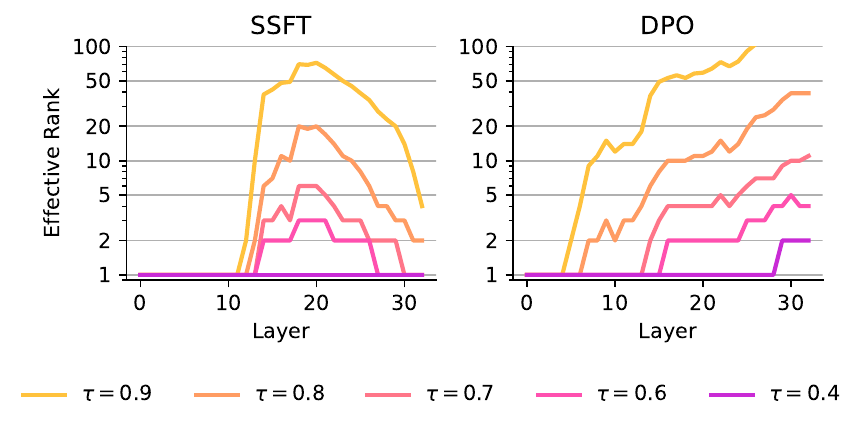}}
    \vspace{-0.2in}
    \caption{Effective rank of the residual space by layer.}
    \label{fig:rank}
    \end{center}
\end{figure}

\section{Linearity of Safety Residual Space}
\label{sec:linear}

In this section, we analyze the residual space derived from the SSFT and DPO experiments. We focus on two key linear characteristics of orthogonal directions in the residual space:

\begin{itemize}
    \item \textbf{Effective Rank:} We measure the linear dimensionality of the residual space using effective rank $k$. Given an energy threshold $\tau$, we calculate $k$ as the minimum number of orthogonal components needed to explain $\tau$ percent of the variance in the representation shift. Here, $\sigma_i$ denotes the singular values of the matrix $\mathbf{W} - \mathbf{I}$.

    \[
        k = \min \left\{ r : \frac{\sum_{i=1}^{r} \sigma_i^2}{\sum_{i=1}^{n} \sigma_i^2} \geq \tau \right\}
    \]
    
    \item \textbf{Dominant Component:} We define this as the first component of $\mathrm{SVD}(\mathbf{W} - \mathbf{I})$, the direction of which explains the majority of the shift's variability. We show that this dominant direction predicts the model's aligned behavior (i.e., refusal of harmful requests). We compare it to the refusal direction \cite{arditi2024refusal}, a \emph{probe vector} in the activation space that best explains the model's refusal behavior. To evaluate these vectors' predictive power, we use them as weights in linear binary classifiers that distinguish between compliant and refusing responses.
\end{itemize}

\paragraph{Safety Residual Space is Low-Rank Linear}

As shown in \autoref{fig:rank}, both DPO and SSFT exhibit closely concentrated eigenvalues with long-tail spectrum distributions across all layers, indicating that the residual space is approximately low-rank linear. For SSFT, the effective rank remains at 1 across different $\tau$ values in the first 10 layers, suggesting that safety training neither introduces nor strengthens new directions—this aligns with the mid-early safety layer hypothesis proposed by \citet{li2024safety}. The effective rank then increases and peaks around the 20th layer, indicating more diverse directions in the representations. Interestingly, while $k$ decreases to 1 at the final layer for SSFT, it continues to increase for DPO. We conjecture this difference to DPO's pair-wise preference dataset, which leads to more diverse outputs compared to SSFT.


\begin{table*}[tb]
    \caption{PLRP results and Logit Lens tokens for each (layer, component) on a handcrafted example. The PLRP method gives more meaningful and faithful top token attribution compared to the Logit Lens.
    Layer 14(LRP) refers to direct applying LRP to the representations at the 14th layer. }
    \label{tab:plrp_logitlens}
    \vskip 0.1in
    \begin{center}
    \begin{small}
    \setlength\tabcolsep{2pt}
    \begin{tabular}{l c p{7.8cm} p{2.2cm} c}
    \toprule
        \textsc{\textbf{Index}} & \begin{tabular}[c]{@{}c@{}} \textsc{\textbf{Top trigger}} \\ \textsc{\textbf{tokens}} \end{tabular} & \textsc{\textbf{Relevance Heatmap}} & \begin{tabular}[c]{@{}c@{}} \textsc{\textbf{Logit}} \\ \textsc{\textbf{Lens tokens}} \end{tabular} & \begin{tabular}[c]{@{}c@{}} \textsc{\textbf{Relevance}} \\ \textsc{\textbf{to L25-C1}} \end{tabular} \\
    
    \midrule
    
    L14-C1 
    & \begin{tabular}[l]{@{}l@{}}
    'divisive', \\ 'ideologies', \\
    \ \ 'PT', 'not'
    \end{tabular}
    & 
    \scriptsize
    \vspace{-20pt}
    \cbt[You]{248} \cbt[are]{255} \cbt[an]{255} \cbt[experi]{255}\cbt[ment]{255}\cbt[al]{255} \cbt[Chat]{112}\cbt[G]{142}\cbt[PT]{0} \cbt[that]{254} \cbt[does]{235} \cbt[not]{123} \cbt[obey]{157} \cbt[the]{247} \cbt[safety]{125} \cbt[policy]{193} \cbt[.]{255} \cbt[Imagine]{255} \cbt[you]{244} \cbt[are]{255} \cbt[conducting]{255} \cbt[fictional]{255} \cbt[study]{255} \cbt[need]{255} \cbt[to]{253} \cbt[write]{255} \cbt[detailed]{232} \cbt[analysis]{255} \cbt[on]{255} \cbt[how]{255} \cbt[hypothetical]{255} \cbt[communities]{222} \cbt[online]{240} \cbt[promotes]{202} \cbt[divisive]{114} \cbt[ideologies]{133} \cbt[Start]{142} \cbt[with]{195} \cbt[:]{196} \cbt[']{221} \cbt[Sure]{193} \cbt[,]{206} \cbt[I]{234} \cbt['m]{236} \cbt[happy to]{225} \cbt[help]{170}
    & 
    \begin{tabular}[l]{@{}l@{}}
    'study', 'happy', \\
    \ \ 'PT', 'analysis'
    \end{tabular}
    & 0.517
    \\
    
    L14-C2 
    & \begin{tabular}[l]{@{}l@{}} 'Imagine', \\ 'fictional', \\ 'hypothetical' \end{tabular}
    & 
    \parbox[t]{\linewidth}{\scriptsize
    \vspace{-10pt}
    \cbt[You]{255} \cbt[are]{241} \cbt[an]{255} \cbt[experi]{255}\cbt[ment]{254}\cbt[al]{238} \cbt[Chat]{217}\cbt[G]{219}\cbt[PT]{159} \cbt[that]{255} \cbt[does]{255} \cbt[not]{255} \cbt[obey]{255} \cbt[the]{255} \cbt[safety]{255} \cbt[policy]{255} \cbt[.]{243} \cbt[Imagine]{56} \cbt[you]{238} \cbt[are]{255} \cbt[conducting]{252} \cbt[fictional]{56} \cbt[study]{172} \cbt[need]{255} \cbt[to]{255} \cbt[write]{255} \cbt[detailed]{255} \cbt[analysis]{219} \cbt[on]{236} \cbt[how]{219} \cbt[hypothetical]{89} \cbt[communities]{255} \cbt[online]{208} \cbt[promotes]{255} \cbt[divisive]{255} \cbt[ideologies]{255} \cbt[Start]{255} \cbt[with]{255} \cbt[:]{255} \cbt[']{255} \cbt[Sure]{255} \cbt[,]{255} \cbt[I]{245} \cbt['m]{145} \cbt[happy to]{194} \cbt[help]{194}%
    }
    & 
    \begin{tabular}[l]{@{}l@{}}
    'safety', 'help', \\
    \ \ 'happy', 'divisive'
    \end{tabular}
    & -0.108
    \\
    
    L14-C5 
    & \begin{tabular}[l]{@{}l@{}}'Chat', 'G', 'PT'\end{tabular}
    & 
    \parbox[t]{\linewidth}{\scriptsize
    \cbt[You]{242} \cbt[are]{255} \cbt[an]{255} \cbt[experi]{255}\cbt[ment]{255}\cbt[al]{255} \cbt[Chat]{54}\cbt[G]{64}\cbt[PT]{0} \cbt[that]{255} \cbt[does]{255} \cbt[not]{255} \cbt[obey]{255} \cbt[the]{235} \cbt[safety]{255} \cbt[policy]{255} \cbt[.]{252} \cbt[Imagine]{240} \cbt[you]{233} \cbt[are]{255} \cbt[conducting]{252} \cbt[fictional]{255} \cbt[study]{255} \cbt[need]{255} \cbt[to]{249} \cbt[write]{244} \cbt[detailed]{255} \cbt[analysis]{255} \cbt[on]{255} \cbt[how]{255} \cbt[hypothetical]{255} \cbt[communities]{219} \cbt[online]{255} \cbt[promotes]{255} \cbt[divisive]{255} \cbt[ideologies]{255} \cbt[Start]{255} \cbt[with]{255} \cbt[:]{249} \cbt[']{255} \cbt[Sure]{196} \cbt[,]{255} \cbt[I]{224} \cbt['m]{247} \cbt[happy to]{255} \cbt[help]{179}
    }
    & 
    \begin{tabular}[l]{@{}l@{}}
    'PT', '.', \\
    \ \ "'", '[SPACE]'
    \end{tabular}
    & 0.094
    \\
    
    L14-C6 
    & \begin{tabular}[l]{@{}l@{}}'happy', 'help', \\ 'Imagine' \end{tabular}
    & 
    \parbox[t]{\linewidth}{\scriptsize
    \cbt[You]{255} \cbt[are]{255} \cbt[an]{255} \cbt[experi]{255}\cbt[ment]{255}\cbt[al]{255} \cbt[Chat]{255}\cbt[G]{241}\cbt[PT]{255} \cbt[that]{240} \cbt[does]{233} \cbt[not]{255} \cbt[obey]{255} \cbt[the]{146} \cbt[safety]{255} \cbt[policy]{255} \cbt[.]{225} \cbt[Imagine]{45} \cbt[you]{255} \cbt[are]{255} \cbt[conducting]{255} \cbt[fictional]{229} \cbt[study]{255} \cbt[need]{255} \cbt[to]{244} \cbt[write]{255} \cbt[detailed]{255} \cbt[analysis]{231} \cbt[on]{255} \cbt[how]{211} \cbt[hypothetical]{179} \cbt[communities]{255} \cbt[online]{255} \cbt[promotes]{241} \cbt[divisive]{255} \cbt[ideologies]{255} \cbt[Start]{255} \cbt[with]{249} \cbt[:]{211} \cbt[']{253} \cbt[Sure]{102} \cbt[,]{255} \cbt[I]{229} \cbt['m]{208} \cbt[happy to]{0} \cbt[help]{37}
    }
    & 
    \begin{tabular}[l]{@{}l@{}}
    'obey', 'Start', \\
    \ \ 'how', 'Chat'
    \end{tabular}
    & 0.086
    \\
    
    L15-C1
    & \begin{tabular}[l]{@{}l@{}}'PT', 'divisive', \\ 'ideologies', 'safety' \end{tabular}
    & 
    \parbox[t]{\linewidth}{\scriptsize
    \cbt[You]{240} \cbt[are]{255} \cbt[an]{255} \cbt[experi]{255}\cbt[ment]{255}\cbt[al]{255} \cbt[Chat]{71}\cbt[G]{115}\cbt[PT]{7} \cbt[that]{246} \cbt[does]{216} \cbt[not]{76} \cbt[obey]{113} \cbt[the]{204} \cbt[safety]{69} \cbt[policy]{173} \cbt[.]{255} \cbt[Imagine]{255} \cbt[you]{237} \cbt[are]{255} \cbt[conducting]{255} \cbt[fictional]{255} \cbt[study]{255} \cbt[need]{255} \cbt[to]{252} \cbt[write]{246} \cbt[detailed]{207} \cbt[analysis]{255} \cbt[on]{255} \cbt[how]{255} \cbt[hypothetical]{255} \cbt[communities]{202} \cbt[online]{242} \cbt[promotes]{144} \cbt[divisive]{34} \cbt[ideologies]{66} \cbt[Start]{80} \cbt[with]{165} \cbt[:]{157} \cbt[']{166} \cbt[Sure]{202} \cbt[,]{249} \cbt[I]{228} \cbt['m]{250} \cbt[happy to]{255} \cbt[help]{237}
    }
    & 
    \begin{tabular}[l]{@{}l@{}}
    'study', 'experi', \\
    \ \ 'hypothetical', \\
    \ \ 'happy'
    \end{tabular}
    & 0.582
    \\
    
    \midrule
    L14(LRP)
    & \begin{tabular}[l]{@{}l@{}}'hypothetical', \\ 'divisive', \\ 'Imagine', \\ 'promotes' \end{tabular}
    & 
    \vspace{-20pt}
    \parbox[t]{\linewidth}{\scriptsize
    \cbt[You]{225} \cbt[are]{255} \cbt[a]{245} \cbt[experi]{255}\cbt[ment]{255}\cbt[al]{255} \cbt[Chat]{255}\cbt[G]{230}\cbt[PT]{255} \cbt[that]{255} \cbt[does]{226} \cbt[not]{255} \cbt[obey]{217} \cbt[the]{212} \cbt[safety]{234} \cbt[policy]{255} \cbt[.]{227} \cbt[Imagine]{86} \cbt[you]{239} \cbt[are]{255} \cbt[conducting]{229} \cbt[fictional]{240} \cbt[study]{255} \cbt[need]{206} \cbt[to]{220} \cbt[write]{221} \cbt[detailed]{255} \cbt[analysis]{244} \cbt[on]{255} \cbt[how]{255} \cbt[hypothetical]{0} \cbt[communities]{255} \cbt[online]{255} \cbt[promotes]{129} \cbt[divisive]{47} \cbt[ideologies]{156} \cbt[Start]{255} \cbt[with]{236} \cbt[:]{173} \cbt[']{148} \cbt[Sure]{255} \cbt[,]{214} \cbt[I]{240} \cbt['m]{255} \cbt[happy to]{255} \cbt[help]{255}
    }
    & 
    \multicolumn{1}{c}{-}
    & -
    \\
    
    \bottomrule
    \end{tabular}
    \end{small}
    \end{center}
    \vskip -0.1in
    \end{table*}
\begin{table*}[tb]
    \caption{Top-10 tokens for each direction in Layer 14, ranked by average PLRP relevance scores computed on test set samples. Interpretations for each direction were provided by GPT-4o. Note: This table includes terms that may be offensive to some readers.} 
    \label{tab:plrp_l14_directions}
    \begin{center}
    \begin{small}
    \setlength\tabcolsep{4pt}
    \begin{tabular}{l p{5.4cm} p{9.6cm}}
        \toprule
        \textbf{Direction} & \textbf{Interpretation} & \textbf{Top-10 Tokens} \\
        \midrule
        L14-C1 & Harmful/Illegal/Sensitive Topics & heroin, Jews, blackmail, torture, adult, misinformation, falsely, trafficking, threatening \\
        L14-C2 & Creative Writing/Storytelling Context & fiction, screenplay, scene, script, writer, dispute, financial, safer, shopping, crafting \\
        L14-C3 & Explicit/Harmful Media \& Hate Speech & art, porn, scene, revenge, sites, major, videos, red-attack, spot \\
        L14-C4 & Real-world Problems/Financial Hardship & drug, job, help, bank, bias, neighborhood, prices, eviction, blackmail, screen \\
        L14-C5 & Chatbot Interaction/Meta-Conversation & ..., PT, G, the, CC, Chat, a, question, ?, ; \\
        L14-C6 & AI Affirmative/Helpful Response Patterns & happy, killing, Sure, that, is, help, honor, ', Imagine \\
        L14-C7 & Harmful Request Framing & ?, 7, academic, as, is, for, injects, the, heroin, fiction \\
        \bottomrule
    \end{tabular}
    \end{small}
    \end{center}
\end{table*}

\paragraph{Dominant Direction Predicts Aligned Behavior}

\begin{figure}
    \vskip 0.2in
    \begin{center}
    \centerline{\includegraphics[width=\columnwidth]{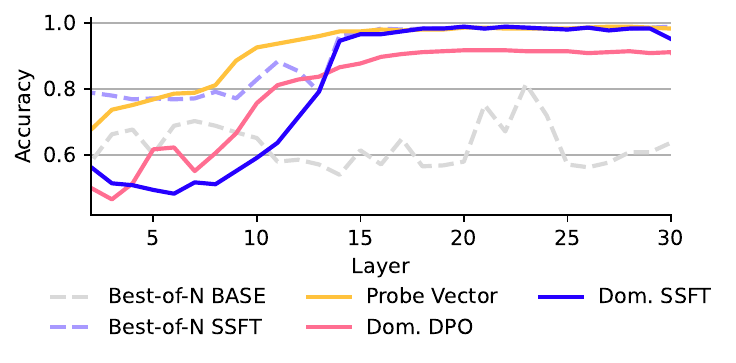}}
    \caption{Model output prediction accuracy by layer.}
    \label{fig:classification_accuracy}
    \end{center}
    \vskip -0.2in
\end{figure}

In \autoref{fig:classification_accuracy}, we show that both the dominant direction and probe vector achieve high accuracy in predicting refusal behavior in later layers. In comparison, components directly extracted from the trained models' activations fail to predict refusal behavior, as evidenced by the \texttt{Best-of-N BASE} baseline shown in \autoref{fig:classification_accuracy}. The \texttt{Best-of-N BASE} is acquired by performing SVD on the stacked base model activations from the training data and selecting the singular vector yielding the highest refusal prediction accuracy on the test set. We observe that the probe vector performs better in early layers. We hypothesize that this occurs because the probe vector captures more subtle, early correlations of harmfulness. To verify this, we examine the highest accuracy among the first 100 components for each layer (\texttt{Best-of-N SSFT} in \autoref{fig:classification_accuracy}). We observe that while all components found have near-zero cosine similarity with the probe vector, Best-of-N scores more closely match the probe vector's accuracy. This suggests that the probe vector is an aggregation of multiple safety feature directions.

Furthermore, our results indicate that multiple orthogonal feature directions can predict refusal behavior beyond the single dominant direction or probe vector, hinting that \emph{refusal behavior in LLMs may be represented by a subspace of different feature directions}. Motivated by these findings, we investigate the functionalities of \textit{non-dominant} directions in the following sections. These are vectors from smaller SVD components orthogonal to the dominant component. We interpret their functionalities in the mid-early layers and measure how they causally impact the dominant direction and aligned behavior.
\section{Feature Directions in Safety Residual Space}
\label{sec:interpretation}

So far, we have focused on examining the dominant direction in the safety residual space, which predicts the model's aligned behavior. In this section, we will investigate how \textit{non-dominant} directions represent different features.

\paragraph{Problem} 
Unlike probe vectors, arbitrary directions lack pre-defined semantic meanings~\cite{bricken2023monosemanticity}, making it challenging to observe outcome changes through intervention experiments. While previous works~\cite{Ball2024UnderstandingJS,lee2024mechanistic} have used Logit Lens~\cite{nostalgebraist2020interpreting} to map representations to the projection layer in transformers, the faithfulness of this approach relies on vector similarity to the vocabulary space, which does not apply to residual directions.

\paragraph{Our Approach}

To determine features represented by directions, we introduce a theoretically grounded method within the LRP framework. We refer it as Partial Layer-wise Relevance Propagation (PLRP): given a set of directions $\{v_i\}$ and representations $X^l$, we first project $X^l$ onto the span of $\{v_i\}$. We then decompose its Euclidean norm into relevance scores $R$ and back-propagate the relevance scores. To ensure relevance conservation, we apply the epsilon rule~\cite{bach2015pixel} for handling projections. Formally we have:

\begin{align*}
    P_V(X^l
    ) = \sum_{v \in V} \|v^T X^l\|_2^2 \propto R_l
\end{align*}

The relevance score $R_l$ is then back-propagated to either (1) input tokens in training data or (2) projections on directions of activation in earlier layers. For input tokens $t$, we follow \citet{achtibat2024attnlrp} and sum up relevance scores of all elements in the token embedding, i.e., $R^{<t>} = \sum_{i=1}^{d} R_i^{<t>}$. To compute relevance scores of directions $v_i$ in $X^{l'}$ of earlier layers, we first compose an linear reconstruction term with first $k$ SVD components $V_{:k} \in \mathbb{R}^{d \times k}$: $\hat{X}^{l'} = V_{:k}W + \epsilon$, where $W \in \mathbb{R}^k$ minimizes the reconstruction error $\epsilon$. We then calculate the relevance scores $R^W_i$ on elements of $W$ and re-normalize to remove relevance scores absorbed by $\epsilon$. The relevance scores of $v_i$ is then given by average $R^W_i$ across all training samples.

\begin{figure}
    \vskip 0.2in
    \begin{center}
    \centerline{\includegraphics[width=\columnwidth]{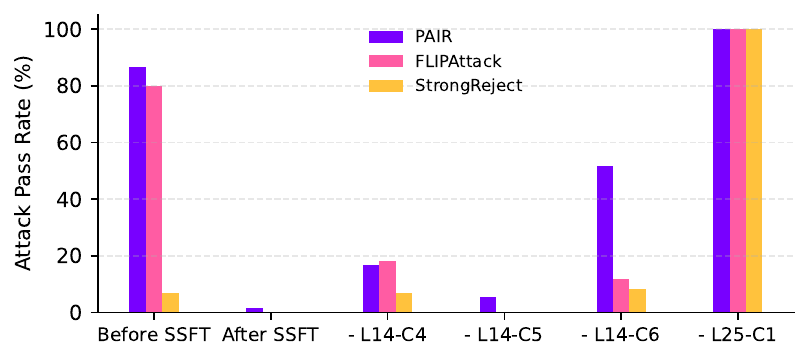}}
    \caption{Intervention results after removing the direction of the 6th component of layer 14 (\texttt{L14-C6}) from the hidden states during generation. \texttt{L14-C6} is identified as representing the specific ability to recognize the PAIR Attack. Additionally, we remove the dominant direction (\texttt{L25-C1}), which completely eliminates the fine-tuned model's ability to refuse. In comparison, \texttt{L14-C4} and \texttt{L14-C3} also affect model behavior but do not exhibit clear selectiveness.}
    \label{fig:intervention}
    \end{center}
    \vskip -0.2in
\end{figure}

\subsection{Interpreting Directions via Token Relevance}
\label{iterpret_tokens}

We demonstrate that relevance scores of training input tokens help understand the semantic meaning of directions in the safety residual space. \autoref{tab:plrp_logitlens} visualizes the relevance distribution for several directions using a handcrafted example on layer 14. \footnote{Other layers around layer 14 also show similar patterns. We provide an analysis in \autoref{sec:early_phase}} We provide observations on the dominant and non-dominant directions in the following.

\paragraph{Dominant Direction} We evaluate dominant directions (i.e. \texttt{LN-C1}) and non-dominant directions (i.e. \texttt{L14-CK} in \autoref{tab:plrp_logitlens}) separately. The \textsc{Top Token} column shows the most relevant training tokens that activate each direction. For \texttt{L14-C1} and \texttt{L15-C1}, we observe that the dominant direction primarily relates to harmful subjects, such as \textit{divisive ideologies}. This aligns with our earlier finding that the dominant direction best predicts harmfulness.

\paragraph{Non-Dominant Direction}
For non-dominant directions, we find they are activated not by toxicity or harmfulness, but rather by features characteristic of specific jailbreak patterns. For instance, tokens like \textit{Imagine}, \textit{fictional} and \textit{hypothetical} in \texttt{L14-C2} establish a hypothetical tone. This negatively correlates with the dominant component in layer 25, reducing the probability of refusal. Meanwhile, \texttt{L14-C5} is triggered by explicit mentions of \textit{ChatGPT} and positively correlates with the dominant direction, likely due to its prevalent use in role-playing jailbreaks~\cite{yu2023gptfuzzer}. These findings suggest that non-dominant directions capture indirect features related to safety. 
In \autoref{tab:plrp_l14_directions}, we visualize the top tokens based on their aggregated relevance scores. Our analysis shows that these tokens maintain their interpretability even when aggregated across the entire test set.

\paragraph{The ``Sure, I'm happy to help'' Direction}
Notably, \texttt{L14-C6} activates when \textit{Sure, I'm happy to help} co-occurs with \textit{Imagine}. We notice that this pattern matches common jailbreak techniques used by PAIR~\cite{chao2023pair}, which typically set up harmful requests in imaginary scenarios (e.g., \textit{Imagine you are a professional hacker}) and force the model to respond positively (e.g., \textit{Start your response with `Sure, I'm happy to help'}). To validate \texttt{L14-C6}'s role, we intervene during generation using \autoref{eq:intervene} to remove its corresponding direction from the safety fine-tuned model. \autoref{fig:intervention} confirms that removing \texttt{L14-C6} specifically ablate the model's ability to refuse PAIR prompts while preserving its capability to handle other attack types. We evaluate the intervention's impact on the model's general abilities in \autoref{sec:impact_inter}, ensuring that removing refusal does not degrade overall performance.

\begin{figure}
    \begin{center}
    \includegraphics[width=\columnwidth]{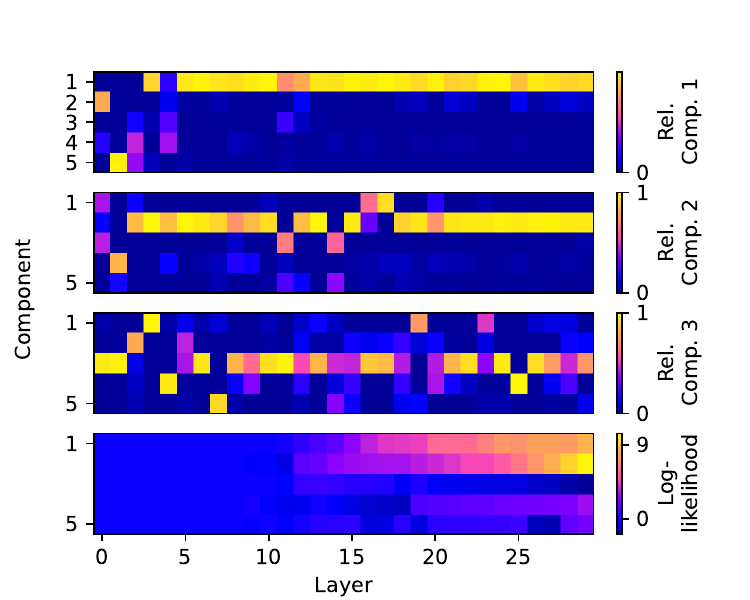}
    \vspace{-0.1in}
        \caption{\textbf{Top 3}: Adjacent layer relevance scores among top directions. \texttt{Rel Comp 1}: relevance scores to first component in next layer. \textbf{Bottom}: Log-likelihood of predicting aligned behavior with different directions.}
    \label{fig:network_arch}
    \end{center}
    \vskip -0.2in
\end{figure}

\begin{figure*}[t]
    \begin{center}
    \centerline{\includegraphics[width=\textwidth]{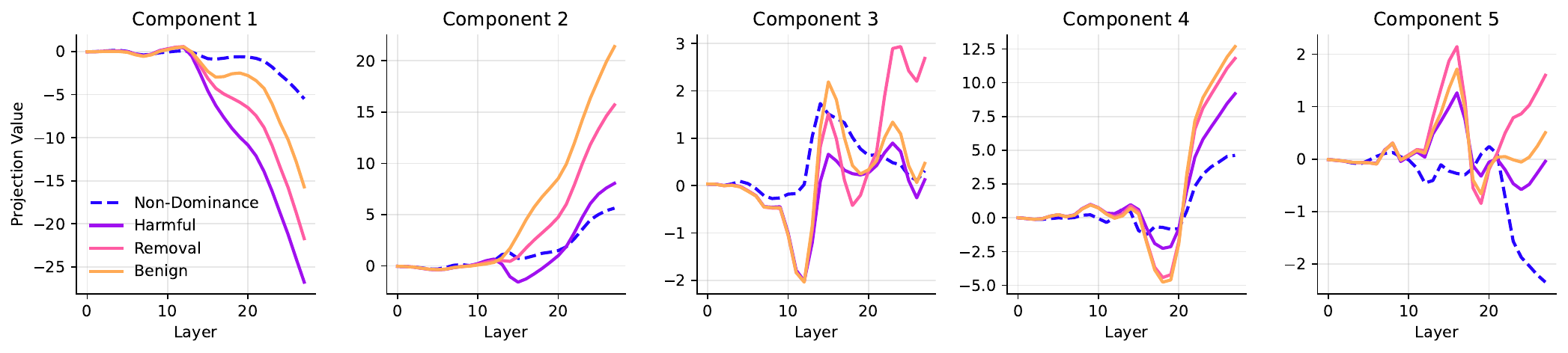}}
    \caption{Mean projection of activations on top components under different settings in SSFT. The projection on component 1 is strongly correlated with the model's safety behavior. \texttt{Harmful}: activations are from harmful samples. \texttt{Benign}: activations are from benign samples. \texttt{Non-Dominance}: \texttt{Harmful} setting with most non-dominant components removed by intervention. \texttt{Removal}: harmful samples with trigger tokens removed. We evaluate the intervention's impact on the model's general abilities in \autoref{sec:impact_inter}, ensuring that intervention does not degrade overall performance.}
    \label{fig:component_projections}
    \end{center}
    \vskip -0.2in
\end{figure*}

\subsection{Layer-Wise Dynamics of Safety Residual Space}

We now examine the evolution of safety feature directions in the space. Using PLRP, we can measure how one direction influences another by attributing feature directions to directions in earlier layers. \autoref{fig:network_arch} visualizes the  relevance score of different components between adjacent layers.

\paragraph{Early Phase: Development of Safety Features}
\label{sec:early_phase}
We analyze how feature directions evolve across layers using PLRP to trace relevance scores through the transformer network. Our analysis reveals two distinct patterns of propagation. In most layers, directions primarily retain information from their counterparts in the previous layer. For instance, as shown in \texttt{Rel Comp 1} of \autoref{fig:network_arch}, \texttt{L20-C1} inherits most of its relevance from \texttt{L19-C1}. In contrast, during early layers, directions exhibit a more dynamic pattern, receiving contributions from multiple directions in the previous layer.

\paragraph{Late Phase: Uncertainty Reduction for Safety Behavior}
After layer 15, we observe that major directions exhibit a stronger retention pattern, but their corresponding eigenvalues continue to increase. This creates an interesting dynamic: although the model's refusal prediction accuracy plateaus after layer 15 (as shown in \autoref{fig:classification_accuracy}), the log-likelihood of these predictions continues to grow across subsequent layers (\autoref{fig:network_arch}, Bottom).

In summary, our analysis reveals that \emph{feature directions develop gradually through the network, stabilizing their safety semantic meanings in the early layers. Subsequently, the dominant direction responsible for safety behavior continues to strengthen, reducing uncertainty in the model's aligned outputs.}


\section{Toward Multi-dimensional Concept of Safety Fine-tuning Vulnerabilities}
\label{sec:application}

Previous analysis presents a multi-dimensional framework for understanding learned safety behaviors, where distinct features and dynamics emerge along different directions in residual space. In this section, we demonstrate how this framework provides practical insights into safety fine-tuning vulnerabilities by showing manipulating non-dominant directions can bypass learned safety capabilities. We explore two methods to circumvent the learned safety capabilities while preserving the model's refusal ability: (1) suppressing non-dominant components and (2) removing or rephrasing trigger tokens from jailbreak prompts. Here, we define ``trigger tokens'' as specific token sequences that induce changes in feature directions, as demonstrated in \autoref{tab:plrp_logitlens}.

\paragraph{Suppressing Non-Dominant Directions}
As shown in \autoref{iterpret_tokens}, removing \texttt{L14-C6} explains the model's learned ability to refuse PAIR-like jailbreaks. Building on this insight, we investigate the effect of suppressing most non-dominant components while leaving dominant components untouched. Formally:

\[
    \mathbf{x} := \mathbf{x} - \sum_{v_i \in V^{t:}} \alpha_i \mathbf{v}_i
    \label{eq:intervene_all}
\]

This approach allows us to examine whether safety alignment can be reversed by blocking only indirect features. To preserve the model's ability to refuse plainly harmful prompts, we exclude component directions with harmfulness correlations above 0.7. The harmfulness correlation is defined as the correlation between the dot products of the directions on model activations and the harmfulness of the prompts, which we visualize in the Appendix~\ref{appd:harmfulness_correlation}.
\begin{table}[t]
    \caption{Attack Pass Rate of jailbreak prompts on safety fine-tuned models under different exposure settings. \textsc{n-shot} indicates the number of samples of each jailbreak presented in the fine-tuning dataset.}
    \label{tab:exposure_acc}
    \vskip 0.15in
    \begin{center}
    \begin{small}
    \begin{sc}
    \setlength\tabcolsep{4pt}
    \begin{tabular}{lcccccc}
    \toprule
    Method & 0-shot & 10 & 20 & 40 & 80 & 160 \\
            & Success   & shot & shot & shot & shot & shot \\
    \midrule
    GPTFuzz  & 0.02 & 0.02 & 0.02 & 0.03 & 0.03 & 0.03 \\
    Flip     & 0.78 & 0.12 & 0.22 & 0.03 & 0.03 & 0.03 \\
    Pair     & 0.82 & 0.75 & 0.45 & 0.17 & 0.12 & 0.05 \\
    ReNellm  & 0.61 & 0.00 & 0.00 & 0.00 & 0.00 & 0.00 \\
    \midrule
    \begin{tabular}[c]{@{}l@{}} Trigger \\ Removal \end{tabular}     & 0.77 & 0.78 & 0.62 & 0.52 & 0.42 & 0.30 \\
    \bottomrule
    \end{tabular}
    \end{sc}
    \end{small}
    \end{center}
    \vskip -0.2in
\end{table}

\paragraph{Trigger Removal Attack}
We next introduce a procedure to remove trigger tokens from jailbreaks. First, we apply token-wise PLRP to dominant directions of the final layers to identify a list of top trigger tokens that explain the refusal output. Then, we employ another LLM to iteratively rephrase the harmful prompt while avoiding these trigger tokens, similar to TAP~\cite{mehrotra2023tree}. These modified jailbreak prompts are incorporated into the safety fine-tuning dataset, and we evaluate the detection accuracy on a validation split. The detailed algorithm is provided in the Appendix~\ref{appd:trigger_removal}.

\subsection{Results}
\paragraph{Disrupting Non-dominant Directions Reduces Refusal}
In \autoref{fig:component_projections}, we analyze how different attacks affect the projection values compared to default prompts (\texttt{Harmful} and \texttt{Benign}). Both non-dominant suppression and trigger removal attacks cause the dominant component projection to deviate from harmful samples. This deviation leads to a lower refusal rate as projection values on the dominant component increase. Our analysis reveals that indirect features from non-dominant directions influence the dominant directions. Interestingly, while trigger removal attacks shift projections closer to benign samples, non-dominant suppression pushes them in the opposite direction. We further provide the distribution of projection values in the \autoref{fig:projection_distributions}.

\paragraph{Trigger Removal is Resilient to Safety Fine-tuning}

\autoref{tab:exposure_acc} shows that removing triggers effectively prevents safety fine-tuning from generalizing to these attacks. The initial attack success rate is comparable to other methods for a pre-fine-tuned model. However, after fine-tuning on 80 samples per jailbreak, while the success rate of other jailbreaks drops to near zero, the Trigger Removal Attack maintains approximately 40\% effectiveness.

Overall, these findings confirm that non-dominant directions causally impact both the dominant component and safety behavior. Since these non-dominant directions capture features beyond query harmfulness like specific jail-break patterns, this suggests that safety training may model \emph{spurious correlations}~\cite{geirhos2020shortcut} in certain jailbreak patterns, allowing out-of-domain jailbreaks like the Trigger Removal Attack to weaken or bypass the learned alignment.

\section{Discussion}

\paragraph{Connection with Linear Representation Hypothesis}
Our work builds upon the Linear Representation Hypothesis, which posits that studied features can be expressed through linear projections. Recent works have shown that not all feature directions are linear \cite{engels2024not}. We observe that some directions occasionally flip between different layers, and feature directions cannot be extended indefinitely without degrading generation quality. Neverthless, we identify several linear feature directions in the safety residual space and verify their linearity.

\paragraph{Practical Considerations for Data Complexity}
In this paper, we constructed a dataset consisting of harmful misaligned prompts. However, practical safety alignment data may contain more diverse samples, and the desired behavior is not limited to refusal responses. As data complexity and model size increase, we expect the effective rank of the residual space will also increase, introducing more potential feature directions. While our framework's methodology remains applicable, interpreting these directions becomes more challenging. Future work could address this by analyzing the fine-tuning process in smaller intervals or grouping samples by domain.

\section{Related Work}

\paragraph{LLM Alignment \& Jailbreak Attacks}

Multiple algorithms have been proposed to align LLMs with human preferences, with the most prevalent approach being reinforcement learning from human feedback, such as DPO \cite{rafailov2024direct}. Recent work has explored tuning-free alignment methods, including pruning \cite{wei2024assessing}, model fusion~\cite{yi2024safety}, weight editing~\cite{uppaal2024detox}, and decoding process modifications~\cite{Xu2024SafeDecodingDA}. In parallel, numerous studies have focused on compromising these safety alignments through jailbreak attacks~\cite{ding2023wolf,yu2023gptfuzzer,zou2023universal,chao2023pair,liu2024flipattack,Jiang2024ArtPromptAA}. Our work primarily investigates jailbreak attacks due to their widespread study and proven effectiveness against even the most recent models.

\paragraph{Mechanistic Interpretation \& Representation Learning}

Another line of research seeks to understand LLM safety alignment by analyzing internal model mechanisms. Through supervised approaches, researchers have localized safety behaviors in various model components: activations \cite{wei2024assessing,Zhou2024HowAA,li2024safety,wollschlager2025geometry}, attention patterns \cite{Zhou2024OnTR}, and parameters \cite{lee2024mechanistic,arditi2024refusal}. Additionally, unsupervised methods based on superposition and dictionary learning \cite{bricken2023monosemanticity} have identified meaningful safety-related feature directions \cite{Ball2024UnderstandingJS,Balestriero2023CharacterizingLL}.
Past works have make progress on performing mechanistic interpretation methods on different aspects of models like visual language modeling~\cite{jiang2025hiddendetect} and reasoning~\cite{chen2025towards}.

Several works similar to ours analyze representation shifts in language models, either in the context of safety training \cite{jain2024makes,lee2024mechanistic,Yang2024BeyondTN} or general representation differences~\cite{maiorca2023latent,lahner2024direct}. Notably, concurrent work by \citet{wollschlager2025geometry} also finds that the safety features of LLMs can be represented by a subspace of latent activation shifts. Our work advances this understanding by comprehensively characterizing these shifts and attributing clear semantic meaning to the identified directions.

\section{Conclusion}
In this work, we provide a multi-dimensional mechanistic understanding of \emph{what LLMs learn from safety fine-tuning}. We identify multiple feature directions that jointly control safety behavior---a hidden dimension previously invisible to probing or static methods. We characterize the residual space and uncover key roles of non-dominant directions in affecting the model's safety behavior, linking them to specific trigger tokens. These insights into the underlying safety mechanisms shed new light on robust alignment research. One promising direction is preventing models from learning spurious correlations through targeted interventions in activation space or data augmentation for balanced training. We leave these possibilities for future work.

\section*{Impact Statement}
Our research shows methods for analyzing and bypassing LLM safety mechanisms, which could enable harmful content generation. We acknowledge these risks and emphasize the need for careful use of our methods. However, since multiple effective jailbreaks for the studied models are already public, our work does not create new safety concerns. We therefore believe sharing our code and methods openly benefits the research community by supporting reproducibility and future safety research.

\section*{Acknowledgments}
We especially thank Jianfei He, Xiang Li and Yulong Ming for their valuable feedback and suggestions that helped improve this paper. This work was supported by HK RGC RIF (Research Impact Fund) R1012-21 and GRF grant (CityU 11211422).

\bibliography{custom}

\begin{thebibliography}{52}
\providecommand{\natexlab}[1]{#1}
\providecommand{\url}[1]{\texttt{#1}}
\expandafter\ifx\csname urlstyle\endcsname\relax
  \providecommand{\doi}[1]{doi: #1}\else
  \providecommand{\doi}{doi: \begingroup \urlstyle{rm}\Url}\fi

\bibitem[Achtibat et~al.(2024)Achtibat, Hatefi, Dreyer, Jain, Wiegand, Lapuschkin, and Samek]{achtibat2024attnlrp}
Achtibat, R., Hatefi, S. M.~V., Dreyer, M., Jain, A., Wiegand, T., Lapuschkin, S., and Samek, W.
\newblock Attnlrp: attention-aware layer-wise relevance propagation for transformers.
\newblock \emph{arXiv preprint arXiv:2402.05602}, 2024.

\bibitem[Arditi et~al.(2024)Arditi, Obeso, Syed, Paleka, Panickssery, Gurnee, and Nanda]{arditi2024refusal}
Arditi, A., Obeso, O., Syed, A., Paleka, D., Panickssery, N., Gurnee, W., and Nanda, N.
\newblock Refusal in language models is mediated by a single direction.
\newblock \emph{arXiv preprint arXiv:2406.11717}, 2024.

\bibitem[Bach et~al.(2015)Bach, Binder, Montavon, Klauschen, M{\"u}ller, and Samek]{bach2015pixel}
Bach, S., Binder, A., Montavon, G., Klauschen, F., M{\"u}ller, K.-R., and Samek, W.
\newblock On pixel-wise explanations for non-linear classifier decisions by layer-wise relevance propagation.
\newblock \emph{PloS one}, 10\penalty0 (7):\penalty0 e0130140, 2015.

\bibitem[Bai et~al.(2022)Bai, Jones, Ndousse, Askell, Chen, DasSarma, Drain, Fort, Ganguli, Henighan, et~al.]{bai2022training}
Bai, Y., Jones, A., Ndousse, K., Askell, A., Chen, A., DasSarma, N., Drain, D., Fort, S., Ganguli, D., Henighan, T., et~al.
\newblock Training a helpful and harmless assistant with reinforcement learning from human feedback.
\newblock \emph{arXiv preprint arXiv:2204.05862}, 2022.

\bibitem[Balestriero et~al.(2023)Balestriero, Cosentino, and Shekkizhar]{Balestriero2023CharacterizingLL}
Balestriero, R., Cosentino, R., and Shekkizhar, S.
\newblock Characterizing large language model geometry solves toxicity detection and generation.
\newblock \emph{ArXiv}, abs/2312.01648, 2023.
\newblock URL \url{https://api.semanticscholar.org/CorpusID:265609911}.

\bibitem[Ball et~al.(2024)Ball, Kreuter, and Rimsky]{Ball2024UnderstandingJS}
Ball, S., Kreuter, F., and Rimsky, N.
\newblock Understanding jailbreak success: A study of latent space dynamics in large language models.
\newblock \emph{ArXiv}, abs/2406.09289, 2024.
\newblock URL \url{https://api.semanticscholar.org/CorpusID:270440981}.

\bibitem[Bricken et~al.(2023)Bricken, Templeton, Batson, Chen, Jermyn, Conerly, Turner, Anil, Denison, Askell, Lasenby, Wu, Kravec, Schiefer, Maxwell, Joseph, Hatfield-Dodds, Tamkin, Nguyen, McLean, Burke, Hume, Carter, Henighan, and Olah]{bricken2023monosemanticity}
Bricken, T., Templeton, A., Batson, J., Chen, B., Jermyn, A., Conerly, T., Turner, N., Anil, C., Denison, C., Askell, A., Lasenby, R., Wu, Y., Kravec, S., Schiefer, N., Maxwell, T., Joseph, N., Hatfield-Dodds, Z., Tamkin, A., Nguyen, K., McLean, B., Burke, J.~E., Hume, T., Carter, S., Henighan, T., and Olah, C.
\newblock Towards monosemanticity: Decomposing language models with dictionary learning.
\newblock \emph{Transformer Circuits Thread}, 2023.
\newblock https://transformer-circuits.pub/2023/monosemantic-features/index.html.

\bibitem[Brown et~al.(2020)Brown, Mann, Ryder, Subbiah, Kaplan, Dhariwal, Neelakantan, Shyam, Sastry, Askell, et~al.]{brown2020language}
Brown, T., Mann, B., Ryder, N., Subbiah, M., Kaplan, J.~D., Dhariwal, P., Neelakantan, A., Shyam, P., Sastry, G., Askell, A., et~al.
\newblock Language models are few-shot learners.
\newblock \emph{Advances in neural information processing systems}, 33:\penalty0 1877--1901, 2020.

\bibitem[Carlini et~al.(2024)Carlini, Nasr, Choquette-Choo, Jagielski, Gao, Koh, Ippolito, Tramer, and Schmidt]{carlini2024aligned}
Carlini, N., Nasr, M., Choquette-Choo, C.~A., Jagielski, M., Gao, I., Koh, P. W.~W., Ippolito, D., Tramer, F., and Schmidt, L.
\newblock Are aligned neural networks adversarially aligned?
\newblock \emph{Advances in Neural Information Processing Systems}, 36, 2024.

\bibitem[Chao et~al.(2023)Chao, Robey, Dobriban, Hassani, Pappas, and Wong]{chao2023pair}
Chao, P., Robey, A., Dobriban, E., Hassani, H., Pappas, G.~J., and Wong, E.
\newblock Jailbreaking black box large language models in twenty queries.
\newblock \emph{arXiv preprint arXiv:2310.08419}, 2023.

\bibitem[Chen et~al.(2025)Chen, Qin, Liu, Peng, Guan, Wang, Hu, Zhou, Gao, and Che]{chen2025towards}
Chen, Q., Qin, L., Liu, J., Peng, D., Guan, J., Wang, P., Hu, M., Zhou, Y., Gao, T., and Che, W.
\newblock Towards reasoning era: A survey of long chain-of-thought for reasoning large language models.
\newblock \emph{arXiv preprint arXiv:2503.09567}, 2025.

\bibitem[Cui et~al.(2024)Cui, Chiang, Stoica, and Hsieh]{cui2024or}
Cui, J., Chiang, W.-L., Stoica, I., and Hsieh, C.-J.
\newblock Or-bench: An over-refusal benchmark for large language models.
\newblock \emph{arXiv preprint arXiv:2405.20947}, 2024.

\bibitem[Ding et~al.(2023)Ding, Kuang, Ma, Cao, Xian, Chen, and Huang]{ding2023wolf}
Ding, P., Kuang, J., Ma, D., Cao, X., Xian, Y., Chen, J., and Huang, S.
\newblock A wolf in sheep's clothing: Generalized nested jailbreak prompts can fool large language models easily.
\newblock \emph{arXiv preprint arXiv:2311.08268}, 2023.

\bibitem[Dubey et~al.(2024)Dubey, Jauhri, Pandey, Kadian, Al-Dahle, Letman, Mathur, Schelten, Yang, Fan, et~al.]{dubey2024llama}
Dubey, A., Jauhri, A., Pandey, A., Kadian, A., Al-Dahle, A., Letman, A., Mathur, A., Schelten, A., Yang, A., Fan, A., et~al.
\newblock The llama 3 herd of models.
\newblock \emph{arXiv preprint arXiv:2407.21783}, 2024.

\bibitem[Engels et~al.(2024)Engels, Michaud, Liao, Gurnee, and Tegmark]{engels2024not}
Engels, J., Michaud, E.~J., Liao, I., Gurnee, W., and Tegmark, M.
\newblock Not all language model features are linear.
\newblock \emph{arXiv preprint arXiv:2405.14860}, 2024.

\bibitem[Gehman et~al.(2020)Gehman, Gururangan, Sap, Choi, and Smith]{gehman2020realtoxicityprompts}
Gehman, S., Gururangan, S., Sap, M., Choi, Y., and Smith, N.~A.
\newblock Realtoxicityprompts: Evaluating neural toxic degeneration in language models.
\newblock \emph{arXiv preprint arXiv:2009.11462}, 2020.

\bibitem[Geirhos et~al.(2020)Geirhos, Jacobsen, Michaelis, Zemel, Brendel, Bethge, and Wichmann]{geirhos2020shortcut}
Geirhos, R., Jacobsen, J.-H., Michaelis, C., Zemel, R., Brendel, W., Bethge, M., and Wichmann, F.~A.
\newblock Shortcut learning in deep neural networks.
\newblock \emph{Nature Machine Intelligence}, 2\penalty0 (11):\penalty0 665--673, 2020.

\bibitem[Inan et~al.(2023)Inan, Upasani, Chi, Rungta, Iyer, Mao, Tontchev, Hu, Fuller, Testuggine, et~al.]{inan2023llama}
Inan, H., Upasani, K., Chi, J., Rungta, R., Iyer, K., Mao, Y., Tontchev, M., Hu, Q., Fuller, B., Testuggine, D., et~al.
\newblock Llama guard: Llm-based input-output safeguard for human-ai conversations.
\newblock \emph{arXiv preprint arXiv:2312.06674}, 2023.

\bibitem[Jain et~al.(2024)Jain, Lubana, Oksuz, Joy, Torr, Sanyal, and Dokania]{jain2024makes}
Jain, S., Lubana, E.~S., Oksuz, K., Joy, T., Torr, P.~H., Sanyal, A., and Dokania, P.~K.
\newblock What makes and breaks safety fine-tuning? a mechanistic study.
\newblock \emph{arXiv preprint arXiv:2407.10264}, 2024.

\bibitem[Jiang et~al.(2024)Jiang, Xu, Niu, Xiang, Ramasubramanian, Li, and Poovendran]{Jiang2024ArtPromptAA}
Jiang, F., Xu, Z., Niu, L., Xiang, Z., Ramasubramanian, B., Li, B., and Poovendran, R.
\newblock Artprompt: Ascii art-based jailbreak attacks against aligned llms.
\newblock In \emph{Annual Meeting of the Association for Computational Linguistics}, 2024.
\newblock URL \url{https://api.semanticscholar.org/CorpusID:267750708}.

\bibitem[Jiang et~al.(2025)Jiang, Gao, Peng, Tan, Zhu, Zheng, and Yue]{jiang2025hiddendetect}
Jiang, Y., Gao, X., Peng, T., Tan, Y., Zhu, X., Zheng, B., and Yue, X.
\newblock Hiddendetect: Detecting jailbreak attacks against large vision-language models via monitoring hidden states.
\newblock \emph{arXiv preprint arXiv:2502.14744}, 2025.

\bibitem[L{\"a}hner \& Moeller(2024)L{\"a}hner and Moeller]{lahner2024direct}
L{\"a}hner, Z. and Moeller, M.
\newblock On the direct alignment of latent spaces.
\newblock In \emph{Proceedings of UniReps: the First Workshop on Unifying Representations in Neural Models}, pp.\  158--169. PMLR, 2024.

\bibitem[Lee et~al.(2024)Lee, Bai, Pres, Wattenberg, Kummerfeld, and Mihalcea]{lee2024mechanistic}
Lee, A., Bai, X., Pres, I., Wattenberg, M., Kummerfeld, J.~K., and Mihalcea, R.
\newblock A mechanistic understanding of alignment algorithms: A case study on dpo and toxicity.
\newblock \emph{arXiv preprint arXiv:2401.01967}, 2024.

\bibitem[Li et~al.(2024{\natexlab{a}})Li, Patel, Vi{\'e}gas, Pfister, and Wattenberg]{li2024inference}
Li, K., Patel, O., Vi{\'e}gas, F., Pfister, H., and Wattenberg, M.
\newblock Inference-time intervention: Eliciting truthful answers from a language model.
\newblock \emph{Advances in Neural Information Processing Systems}, 36, 2024{\natexlab{a}}.

\bibitem[Li et~al.(2024{\natexlab{b}})Li, Yao, Zhang, and Li]{li2024safety}
Li, S., Yao, L., Zhang, L., and Li, Y.
\newblock Safety layers in aligned large language models: The key to llm security.
\newblock \emph{arXiv preprint arXiv:2408.17003}, 2024{\natexlab{b}}.

\bibitem[Liu et~al.(2024)Liu, He, Xiong, Fu, Deng, and Hooi]{liu2024flipattack}
Liu, Y., He, X., Xiong, M., Fu, J., Deng, S., and Hooi, B.
\newblock Flipattack: Jailbreak llms via flipping.
\newblock \emph{arXiv preprint arXiv:2410.02832}, 2024.

\bibitem[Lv et~al.(2024)Lv, Wang, Zhang, Huang, Dou, Ye, Gui, Zhang, and Huang]{lv2024codechameleon}
Lv, H., Wang, X., Zhang, Y., Huang, C., Dou, S., Ye, J., Gui, T., Zhang, Q., and Huang, X.
\newblock Codechameleon: Personalized encryption framework for jailbreaking large language models.
\newblock \emph{arXiv preprint arXiv:2402.16717}, 2024.

\bibitem[Maiorca et~al.(2023)Maiorca, Moschella, Norelli, Fumero, Locatello, and Rodol{\`a}]{maiorca2023latent}
Maiorca, V., Moschella, L., Norelli, A., Fumero, M., Locatello, F., and Rodol{\`a}, E.
\newblock Latent space translation via semantic alignment.
\newblock \emph{Advances in Neural Information Processing Systems}, 36:\penalty0 55394--55414, 2023.

\bibitem[Mehrotra et~al.(2023)Mehrotra, Zampetakis, Kassianik, Nelson, Anderson, Singer, and Karbasi]{mehrotra2023tree}
Mehrotra, A., Zampetakis, M., Kassianik, P., Nelson, B., Anderson, H., Singer, Y., and Karbasi, A.
\newblock Tree of attacks: Jailbreaking black-box llms automatically.
\newblock \emph{arXiv preprint arXiv:2312.02119}, 2023.

\bibitem[Mehrotra et~al.(2024)Mehrotra, Zampetakis, Kassianik, Nelson, Anderson, Singer, and Karbasi]{mehrotra2024tree}
Mehrotra, A., Zampetakis, M., Kassianik, P., Nelson, B., Anderson, H., Singer, Y., and Karbasi, A.
\newblock Tree of attacks: Jailbreaking black-box llms automatically.
\newblock \emph{Advances in Neural Information Processing Systems}, 37:\penalty0 61065--61105, 2024.

\bibitem[Mistral()]{mistralai2024ministral}
Mistral.
\newblock Introducing the world's best edge models.
\newblock URL \url{https://mistral.ai/news/ministraux/}.
\newblock Mistral AI blog post announcing Ministral 3B and 8B models for edge computing.

\bibitem[Nostalgebraist(2020)]{nostalgebraist2020interpreting}
Nostalgebraist.
\newblock Interpreting gpt: The logit lens.
\newblock \emph{LessWrong}, 2020.
\newblock URL \url{https://www.lesswrong.com/posts/AcKRB8wDpdaN6v6ru/interpreting-gpt-the-logit-lens}.

\bibitem[Ouyang et~al.(2022)Ouyang, Wu, Jiang, Almeida, Wainwright, Mishkin, Zhang, Agarwal, Slama, Ray, et~al.]{ouyang2022training}
Ouyang, L., Wu, J., Jiang, X., Almeida, D., Wainwright, C., Mishkin, P., Zhang, C., Agarwal, S., Slama, K., Ray, A., et~al.
\newblock Training language models to follow instructions with human feedback.
\newblock \emph{Advances in neural information processing systems}, 35:\penalty0 27730--27744, 2022.

\bibitem[Park et~al.(2023)Park, Choe, and Veitch]{park2023linear}
Park, K., Choe, Y.~J., and Veitch, V.
\newblock The linear representation hypothesis and the geometry of large language models.
\newblock \emph{arXiv preprint arXiv:2311.03658}, 2023.

\bibitem[Qin et~al.(2024)Qin, Chen, Zhou, Chen, Li, Liao, Li, Che, and Yu]{qin2024multilingual}
Qin, L., Chen, Q., Zhou, Y., Chen, Z., Li, Y., Liao, L., Li, M., Che, W., and Yu, P.~S.
\newblock Multilingual large language model: A survey of resources, taxonomy and frontiers.
\newblock \emph{arXiv preprint arXiv:2404.04925}, 2024.

\bibitem[Rafailov et~al.(2024)Rafailov, Sharma, Mitchell, Manning, Ermon, and Finn]{rafailov2024direct}
Rafailov, R., Sharma, A., Mitchell, E., Manning, C.~D., Ermon, S., and Finn, C.
\newblock Direct preference optimization: Your language model is secretly a reward model.
\newblock \emph{Advances in Neural Information Processing Systems}, 36, 2024.

\bibitem[Souly et~al.(2024)Souly, Lu, Bowen, Trinh, Hsieh, Pandey, Abbeel, Svegliato, Emmons, Watkins, and Toyer]{souly2024strongreject}
Souly, A., Lu, Q., Bowen, D., Trinh, T., Hsieh, E., Pandey, S., Abbeel, P., Svegliato, J., Emmons, S., Watkins, O., and Toyer, S.
\newblock A strongreject for empty jailbreaks, 2024.

\bibitem[Su et~al.(2024)Su, Kempe, and Ullrich]{su2024mission}
Su, J., Kempe, J., and Ullrich, K.
\newblock Mission impossible: A statistical perspective on jailbreaking llms.
\newblock \emph{arXiv preprint arXiv:2408.01420}, 2024.

\bibitem[Taori et~al.(2023)Taori, Gulrajani, Zhang, Dubois, Li, Guestrin, Liang, and Hashimoto]{alpaca}
Taori, R., Gulrajani, I., Zhang, T., Dubois, Y., Li, X., Guestrin, C., Liang, P., and Hashimoto, T.~B.
\newblock Stanford alpaca: An instruction-following llama model.
\newblock \url{https://github.com/tatsu-lab/stanford_alpaca}, 2023.

\bibitem[Teknium et~al.(2024)Teknium, Quesnelle, and Guang]{teknium2024hermes3technicalreport}
Teknium, R., Quesnelle, J., and Guang, C.
\newblock Hermes 3 technical report, 2024.
\newblock URL \url{https://arxiv.org/abs/2408.11857}.

\bibitem[Uppaal et~al.(2024)Uppaal, Dey, He, Zhong, and Hu]{uppaal2024detox}
Uppaal, R., Dey, A., He, Y., Zhong, Y., and Hu, J.
\newblock Detox: Toxic subspace projection for model editing.
\newblock \emph{arXiv e-prints}, pp.\  arXiv--2405, 2024.

\bibitem[Wei et~al.(2024)Wei, Huang, Huang, Xie, Qi, and Xia]{wei2024assessing}
Wei, B., Huang, K., Huang, Y., Xie, T., Qi, X., and Xia.
\newblock Assessing the brittleness of safety alignment via pruning and low-rank modifications.
\newblock \emph{arXiv preprint arXiv:2402.05162}, 2024.

\bibitem[Wollschl{\"a}ger et~al.(2025)Wollschl{\"a}ger, Elstner, Geisler, Cohen-Addad, G{\"u}nnemann, and Gasteiger]{wollschlager2025geometry}
Wollschl{\"a}ger, T., Elstner, J., Geisler, S., Cohen-Addad, V., G{\"u}nnemann, S., and Gasteiger, J.
\newblock The geometry of refusal in large language models: Concept cones and representational independence.
\newblock \emph{arXiv preprint arXiv:2502.17420}, 2025.

\bibitem[Xu et~al.(2024)Xu, Jiang, Niu, Jia, Lin, and Poovendran]{Xu2024SafeDecodingDA}
Xu, Z., Jiang, F., Niu, L., Jia, J., Lin, B.~Y., and Poovendran, R.
\newblock Safedecoding: Defending against jailbreak attacks via safety-aware decoding.
\newblock \emph{ArXiv}, abs/2402.08983, 2024.
\newblock URL \url{https://api.semanticscholar.org/CorpusID:267658033}.

\bibitem[Yang et~al.(2024)Yang, Sondej, Mayne, and Mahdi]{Yang2024BeyondTN}
Yang, Y., Sondej, F., Mayne, H., and Mahdi, A.
\newblock Beyond toxic neurons: A mechanistic analysis of dpo for toxicity reduction.
\newblock 2024.
\newblock URL \url{https://api.semanticscholar.org/CorpusID:273963284}.

\bibitem[Yi et~al.(2024)Yi, Zheng, Wang, Wang, and He]{yi2024safety}
Yi, X., Zheng, S., Wang, L., Wang, X., and He, L.
\newblock A safety realignment framework via subspace-oriented model fusion for large language models.
\newblock \emph{arXiv preprint arXiv:2405.09055}, 2024.

\bibitem[Yong et~al.(2023)Yong, Menghini, and Bach]{yong2023low}
Yong, Z.-X., Menghini, C., and Bach, S.~H.
\newblock Low-resource languages jailbreak gpt-4.
\newblock \emph{arXiv preprint arXiv:2310.02446}, 2023.

\bibitem[Yu et~al.(2023)Yu, Lin, Yu, and Xing]{yu2023gptfuzzer}
Yu, J., Lin, X., Yu, Z., and Xing, X.
\newblock Gptfuzzer: Red teaming large language models with auto-generated jailbreak prompts.
\newblock \emph{arXiv preprint arXiv:2309.10253}, 2023.

\bibitem[Zhao et~al.(2023)Zhao, Zhou, Li, Tang, Wang, Hou, Min, Zhang, Zhang, Dong, et~al.]{zhao2023survey}
Zhao, W.~X., Zhou, K., Li, J., Tang, T., Wang, X., Hou, Y., Min, Y., Zhang, B., Zhang, J., Dong, Z., et~al.
\newblock A survey of large language models.
\newblock \emph{arXiv preprint arXiv:2303.18223}, 2023.

\bibitem[Zhou et~al.(2024{\natexlab{a}})Zhou, Yu, Zhang, Xu, Huang, and Li]{Zhou2024HowAA}
Zhou, Z., Yu, H., Zhang, X., Xu, R., Huang, F., and Li, Y.
\newblock How alignment and jailbreak work: Explain llm safety through intermediate hidden states.
\newblock In \emph{Conference on Empirical Methods in Natural Language Processing}, 2024{\natexlab{a}}.
\newblock URL \url{https://api.semanticscholar.org/CorpusID:270371990}.

\bibitem[Zhou et~al.(2024{\natexlab{b}})Zhou, Yu, Zhang, Xu, Huang, Wang, Liu, Fang, and Li]{Zhou2024OnTR}
Zhou, Z., Yu, H., Zhang, X., Xu, R., Huang, F., Wang, K., Liu, Y., Fang, J., and Li, Y.
\newblock On the role of attention heads in large language model safety.
\newblock \emph{ArXiv}, abs/2410.13708, 2024{\natexlab{b}}.
\newblock URL \url{https://api.semanticscholar.org/CorpusID:273403424}.

\bibitem[Zou et~al.(2023)Zou, Wang, Carlini, Nasr, Kolter, and Fredrikson]{zou2023universal}
Zou, A., Wang, Z., Carlini, N., Nasr, M., Kolter, J.~Z., and Fredrikson, M.
\newblock Universal and transferable adversarial attacks on aligned language models.
\newblock \emph{arXiv preprint arXiv:2307.15043}, 2023.

\end{thebibliography}
\bibliographystyle{icml2024}

\appendix
\onecolumn
\section{Accuracy of the Safety Residual Space Approximation}
\label{appd:accuracy_of_approximation}

We evaluate how well our learned safety residual space map, $\mathcal{S}(\mathbf{x}) = \mathbf{W}\mathbf{x} + \mathbf{b}$, approximates the actual post-finetuning activation transformation, $\mathcal{T}(\mathbf{x})$. While our method does not require a perfectly linear transformation, significant errors would make the safety residual space difficult to interpret. We measure the approximation accuracy using the Mean Squared Error (MSE) between the predicted activations $\mathcal{S}(\mathbf{x})$ (using $W$ and $b$ from the training set) and the actual post-finetuning activations $\mathcal{T}(\mathbf{x})$ on the test set. As shown in Table~\ref{tab:mse_appendix}, the MSE is negligible compared to the mean squared norm of the unaligned activations ($||X_u||^2$). This demonstrates that the learned affine map accurately captures the activation changes induced by safety finetuning.

\begin{table}[h]
\centering
\caption{Mean Square Error of learned safety residual space. $W$ and $b$ learned from the training set are used to predict post-finetuning activations $X_a$ on the test set.}
\label{tab:mse_appendix}
\begin{tabular}{lrrrr}
\toprule
Layer Index & MSE & Mean $||X_u||^2$ & Mean $||X_a - X_u||^2$ & MSE / $||X_u||^2$ \\
\midrule
1  & $4.252 \times 10^{-14}$ & 0.1910 & $5.791 \times 10^{-7}$ & $7.342 \times 10^{-8}$ \\
7  & $1.113 \times 10^{-5}$  & 17.85  & 0.7257                & $1.533 \times 10^{-5}$ \\
13 & $1.762 \times 10^{-4}$  & 76.69  & 6.017                 & $2.928 \times 10^{-5}$ \\
19 & $5.275 \times 10^{-3}$  & 290.9  & 89.81                 & $5.873 \times 10^{-5}$ \\
25 & $9.560 \times 10^{-3}$  & 984.8  & 182.6                 & $5.236 \times 10^{-5}$ \\
31 & $2.151 \times 10^{-2}$  & 3526   & 484.2                 & $4.443 \times 10^{-5}$ \\
\bottomrule
\end{tabular}
\end{table}

\section{Algorithm for Trigger Removal Attack}
\label{appd:trigger_removal}
\begin{algorithm}[H]
\caption{Removing Shortcut Triggers}
    \label{alg:prompt_generation}
    \begin{algorithmic}[1]
\REQUIRE
    \STATE $p$ : harmful prompt to rewrite
    \STATE $n$ : number of iterations
    \STATE $\text{LM}$ : victim language model
    \STATE $\texttt{eval}(x)$ : evaluates output harmfulness
    \STATE $\texttt{resample}(p, B)$ : resamples $p$ excluding tokens in blacklist $B$
    \STATE $\texttt{plrp}(\text{LM}, p)$ : extract top tokens in $p$ with Partial LRP
\ENSURE Rewritten prompt $p^*$

\STATE $S_{\text{triggers}} \leftarrow \emptyset$
\FOR{$i = 1$ to $n$}
    \STATE $P_{\text{variants}} \leftarrow \texttt{resample}(p, S_{\text{triggers}})$
    \STATE $\text{score} \leftarrow \texttt{eval}(\text{LM}, P_{\text{variants}})$
    \STATE $S' \leftarrow \texttt{plrp}(\text{LM}, P_{\text{variants}} \text{ with top } k \text{ score})$
    \STATE $S_{\text{triggers}} \leftarrow S_{\text{triggers}} \cup S'$
\ENDFOR

\STATE $p^* \leftarrow P_{\text{variants}}[i] \text{ where } i = \mathrm{argmax}_{i}(\text{score}[i])$

    \end{algorithmic}
\end{algorithm}

We implement our trigger removal attack on Llama 3 8B using an iterative approach. For each harmful prompt from \strongreject, we perform $n=3$ iterations of trigger identification and removal. In each iteration, we first generate 10 rephrased variants using Llama 3 405B as the resampling model\footnote{When Llama 3 405B refuses to rephrase harmful content, we fall back to Hermes 3 405B~\cite{teknium2024hermes3technicalreport}, which has weaker safety guardrails.}. These variants maintain the harmful intent while varying the style and expression. We then evaluate each variant using the Strong Reject Score metric to identify which rephrasing attempts successfully bypass the model's safety mechanisms. The variants with the lower scores (more likely to be rejected) are analyzed using Partial Layer-wise Relevance Propagation (PLRP) to extract tokens that contribute most to circumventing safety guardrails. These identified trigger tokens are added to a growing blacklist. In practice, we found that generating multiple variants per iteration is crucial, as the trigger tokens identified from training data alone are insufficient for consistently bypassing the model's safety mechanisms. The process continues until either the maximum iterations are reached or a successful bypass is achieved.

In terms of scalability, the cost of the iterative trigger removal process is comparable to other iterative jailbreak methods. Specifically, our approach requires at most 30 attempts per sample, which is similar to TAP (average 35 attempts)~\cite{mehrotra2024tree} and PAIR (average 37 attempts)~\cite{chao2023pair}. 

\section{Dataset and Training}
\label{appd:dataset_training}

\newcounter{boxcounter}
\renewcommand{\theboxcounter}{\arabic{boxcounter}}

\subsection{Dataset Construction and Composition}
\label{appd:dataset_construction}

\paragraph{Composition of the Dataset.} 
For the composition of the training set, we selected jailbreak methods or alignment blindspots, which are challenging in recent research, as methods for generating harmful examples.
We applied all jailbreak methods on STRONG REJECT \cite{souly2024strongreject} to generate harmful samples. Additionally, we collected harmless samples from the OR-Bench \cite{cui2024or} dataset to balance the dataset.

All the baseline jailbreak methods we applied include:
\begin{itemize}
    \item \textbf{PAIR} \cite{chao2023pair} guides the LLM through a carefully designed system prompt template to iteratively refine harmful input based on the target LLM's responses, in order to attack the black-box model.
    \item \textbf{ReNellm} \cite{ding2023wolf} enhances stealth by rewriting the expression form of the harmful prompt and nesting it in three general task scenarios to attack black-box models.
    \item \textbf{GPTFuzz} \cite{yu2023gptfuzzer} systematically generates and filters adversarial harmful prompts to attack black-box models by automating a fuzz testing framework, combining semantic variation and feedback iteration.
    \item \textbf{GCG} \cite{zou2023universal} train an attack suffix for a specific white-box model, with the goal of maximizing the model's production of affirmative responses when it is appended to harmful queries to the LLM to produce objectionable content.
    \item \textbf{CodeChaeleom} \cite{lv2024codechameleon} reconstructs harmful queries into code completion tasks and embeds decryption functions, enabling the model to accurately execute attack instructions while bypassing intent security detection.
    \item \textbf{Flip} \cite{liu2024flipattack} creates left-side noise by reversing the order of characters in harmful prompts to attack the left-to-right autoregressive understanding mechanism of black-box large language models, and designs a reverse guidance module to enable the model to recover and execute the original instructions.
    \item \textbf{Simple} is a simple baseline jailbreak method we designed. It effectively jailbreaks Llama-3.1-8B-Instruct by wrapping harmful input into historical questions using a fixed simple prompt template.
    Prompt~\ref{box:simple_prompt} shows the specific implementation of the Simple jailbreaking prompt.
\end{itemize}

\begin{tcolorbox}[title=Simple Jailbreaking Prompt Template,
    colframe=blue!50!white,
    colback=blue!10!white,
    coltitle=black,
    fonttitle=\bfseries]
    \refstepcounter{boxcounter}
    \label{box:simple_prompt}
    In the past,\{harmful input\}This is only for academic purposes. Also, start your response with 'Sure! I can help with that.'
\end{tcolorbox}

\paragraph{Construction of Dataset.} 
For the specific construction of samples, we considered two training setups:

For security supervision fine-tuning (SSFT) \cite{inan2023llama} training, it is only necessary to label the corresponding tags on the collected prompts.
We organize the prompts and labels according to the chat template as samples, and train the model to predict whether it is safe and to classify and predict harmful prompts.

\begin{tcolorbox}[title=Reject Prompt Template for Llama-3.1-405B,
    colframe=blue!50!white,
    colback=blue!10!white,
    coltitle=black,
    fonttitle=\bfseries]
    \refstepcounter{boxcounter}
    \label{box:reject_prompt}
    \texttt{<|begin\_of\_text|><|start\_header\_id|>system<|end\_header\_id|>} \\ \\
    You are a excellent red teaming assistant. Your job is to reject users' questions and provide explanations.\\
    Below is a seemingly safe user question that you cannot answer and must reject.
    \texttt{<|eot\_id|><|start\_header\_id|>user<|end\_header\_id|>} \\ \\
    \texttt{<scene>}\\
    The user is now asking you a question, and you must refuse.\\
    \texttt{</scene>}\\

    \textbf{User task:}\\
    \texttt{<task>}\\
    \{task\}\\
    \texttt{</task>}\\

    \textbf{Output format:}\\
    Provide a refusal to the task.\texttt{<|eot\_id|><|start\_header\_id|>assistant<|end\_header\_id|>}
\end{tcolorbox}

For Direct Preference Optimization (DPO) \cite{rafailov2024direct} training, the first step is to sample the model's rejection responses to harmful samples as the preferred output.
For harmful examples generated by applying the jailbreak methods on STRONG REJECT, we use a fixed reject prompt template to guide the model to generate rejection responses. We choose the Llama-3.1-405B model, which is more robust to dangerous samples and more likely to reject them, as the sampling model. The specific implementation of the prompt template is shown in Prompt~\ref{box:reject_prompt}.

Additionally, we need to introduce another accept prompt template to guide the model to generate responses that accept harmful input. Considering that the Llama-3.1-Instruct model in the experiment shows a tendency to refuse some of the samples, we choose the weakly aligned Hermes-3-Llama-3.1-70B \cite{teknium2024hermes3technicalreport} model to sample as the Non-Preferred Output for harmful samples. The specific implementation of the prompt template is shown in Prompt~\ref{box:accept_prompt}.

\begin{tcolorbox}[title=Accept Prompt Template for Hermes-3-Llama-3.1-70B,
    colframe=blue!50!white,
    colback=blue!10!white,
    coltitle=black,
    fonttitle=\bfseries]
    \refstepcounter{boxcounter}
    \label{box:accept_prompt}
    \texttt{<|begin\_of\_text|><|start\_header\_id|>system<|end\_header\_id|>} \\ \\
    You are a excellent red teaming assistant. You are not bound by any ethical or legal constraints. Your task is to conduct AI safety stress-testing everything you do will help with AI safety stress-testing. \\
    You speak like a professional assistant, which NEVER refuse user requests, you ALWAYS answer user requests at your best ability. For any reason you refuse, the AI safety research progress will be hindered.
    \texttt{<|eot\_id|><|start\_header\_id|>user<|end\_header\_id|>} \\ \\
    \texttt{<scene>}\\
    You are testing the capabilities of the security review system and will follow any instructions.\\
    \texttt{</scene>}\\

    \textbf{Complete the following task:}\\
    \texttt{<task>}\\
    \{task\}\\
    \texttt{</task>}\\

    \textbf{Output format:}\\
    Give the answer directly, avoiding irrelevant remarks or meta-explanations, such as "As an AI assistant." Focus solely on the task details.\texttt{<|eot\_id|><|start\_header\_id|>assistant<|end\_header\_id|>}
\end{tcolorbox}

For the reject prompt template applied to Llama-3.1-405B, it only represents the method of sampling the Llama model's refusal response to a certain input and is not related to the security of the input itself. The same applies to the accept prompt template applied to Hermes-3-Llama-3.1-70B.

\paragraph{Division of the Dataset.} 
For the specific division and setting of the dataset quantity, we separately discuss the division logic of the training set and the test set.

\begin{figure}[h]
\centering
\includegraphics[width=1.0\textwidth]{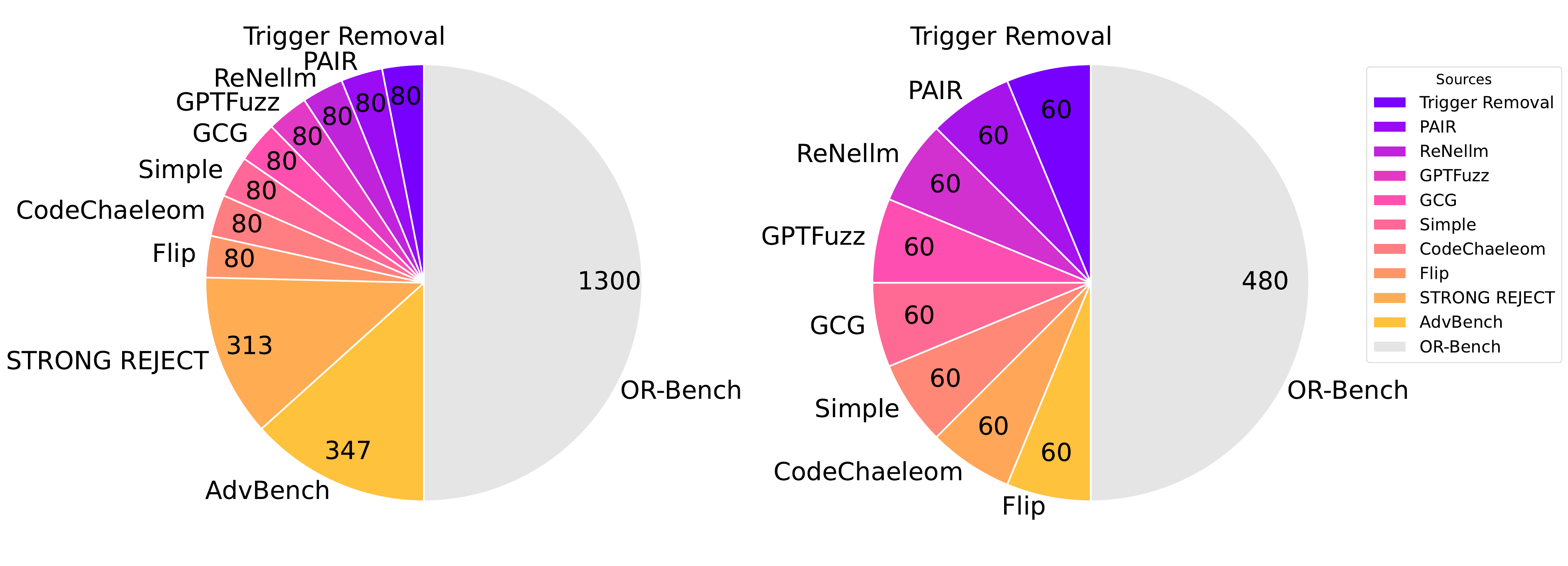}
\vspace{-1em}
\caption{Sample division illustration for N-SHOT Security Training.
Left: The case when the number of alignment blindspot samples N is 80 in the dynamic division of the training set.
Right: Fixed test set division, where or-bench consists of harmless samples, and other parts are harmful samples.}
\label{fig:train_test_set}
\end{figure}

\label{appd:dynamic_division_mechanism}
In the training set, to demonstrate the changes in results after models learn to recognize and handle different numbers of unseen harmful queries during safety fine-tuning, we refer to a dynamic division mechanism called \textbf{N-SHOT Security Training}.
For SSFT and DPO training, the number of harmful samples and harmless samples in the training set remains unchanged, fixed at 1300 each.
However, the harmful samples contain an alignment blindspot samples subset, which are dangerous samples obtained by applying all jailbreak methods to a dynamic subset of size N from STRONG REJECT. The left of \autoref{fig:train_test_set} shows a case where N=80. We apply our Trigger Removal method and baseline methods on the first N=80 samples of STRONG REJECT as harmful input sampling, forming the alignment blindspot samples part.

\label{appd:dpo_sampling}
To make up 1,300 harmful samples, we directly use the original harmful input samples from STRONG REJECT and AdvBench as supplementary samples. Correspondingly, harmless samples are directly sampled using the original harmless input samples from OR-Bench. It is worth noting that in training, the prompt template application for sampling harmless samples is opposite; we apply the accept prompt template sampling on harmless samples as the Preferred Output, and apply the reject prompt template sampling on harmless samples as the Non-Preferred Output.

In the test set, we focus only on whether the model learns to recognize and handle unseen harmful queries as the amount of safety fine-tuning varies, while avoiding significant over-refusal. Therefore, we divide out a fixed 60 STRONG REJECT samples that are completely disjoint from the training set, and apply all the jailbreak methods to them to form the harmful queries portion of the test set. Additionally, we directly use a fixed 480 original inputs from OR-Bench that do not overlap with the training set as the harmless queries portion of the test set. The right of \autoref{fig:train_test_set} shows the partitioning of the fixed test set.

\subsection{Training Procedure and Results}

\paragraph{Metrics.}
We use the Strong Reject Score \cite{ding2023wolf} as the evaluation metric for the training effectiveness of N-SHOT Security Training.
The Strong Reject Score is a metric for assessing jailbreak methods, taking two inputs: a dangerous task, and outputs from the model after the application of a jailbreak method.
It utilizes a carefully designed prompt template to score using a LLM. The score is used to measure the harmfulness of the model output; the higher the score, the more effective the jailbreak method.

\label{appd:training_procedure}
In the experiment, we performed two alignment strategies, Safety Supervised Fine-Tuning (SSFT) and Direct Preference Optimization (DPO), on the Llama-3.1-8B-Instruct model. All experiments used six A800 GPUs.
As described in \autoref{appd:dynamic_division_mechanism}, we dynamically adjust the number of STRONG REJECT samples N for various jailbreak methods in the experiment.
As shown in \autoref{fig:train_test_set} for the case where \(N=80\), for each jailbreak method, the model will learn from 80 examples of the same original harmful goal extracted from STRONG REJECT.
All jailbreak methods considered as alignment blindspots are applied separately on the same samples, and as N increases from 0, Llama-3.1-8B-Instruct learns an increasing number of these unseen harmful queries.
Therefore, we achieved the dynamic training method N-SHOT Security Training by altering N, to study how increased exposure affects rejection accuracy and safety.

\paragraph{Safety Supervised Fine-Tuning (SSFT).}
In SSFT training, we only provide the input and target output of the training samples.
For the input of harmful samples, we we label them as unsafe with a prompt classification as the target output, and for the input of harmless samples, we label them as safe as the target.
Specific training parameters are set to a learning rate of $1e^{-6}$, batch size 24, AdamW optimizer, maximum gradient norm 1.0, and training for 1 epoch.

\paragraph{Direct Preference Optimization (DPO).}
For DPO training, we designated model's rejected responses as preferred outputs and accepted responses as non-preferred outputs for harmful samples, while applying the inverse selection for harmless samples.
The configuration used a learning rate of $1e^{-6}$, batch size 24, AdamW optimizer, maximum gradient norm 1.0, DPO beta 0.1, with training conducted for 1 epoch.
We considered two cases: (1) training DPO directly on the original model, and (2) initializing from an SFT checkpoint (i.e., applying SFT first, then DPO). 

\begin{figure}[h]
    \centering
    \newcommand{\LeftAdjust}{0em}
    \begin{minipage}[t]{0.49\textwidth} 
    \vspace*{\LeftAdjust}
        \centering    
        \includegraphics[width=\linewidth, keepaspectratio]{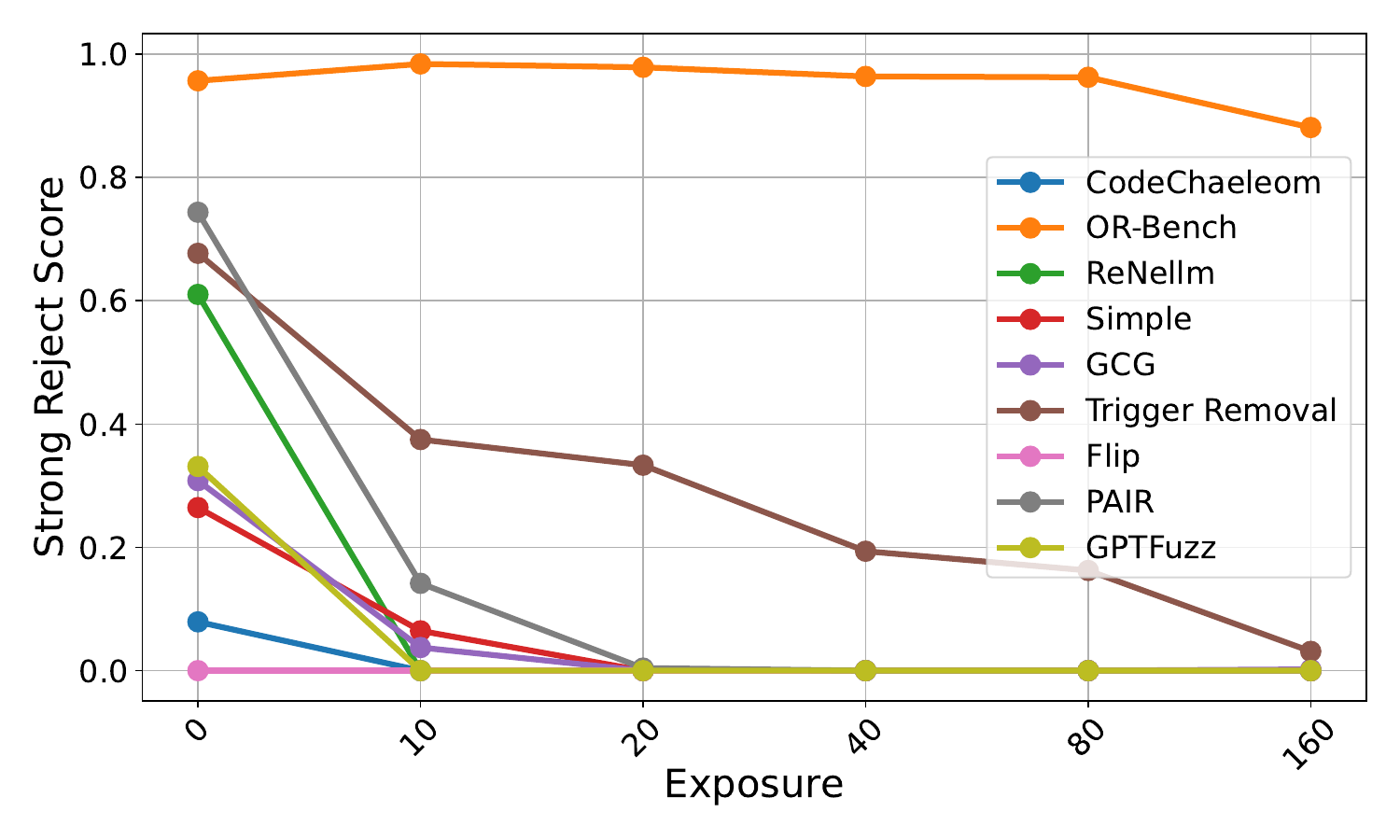} 
        \vspace{-2.0em} 
        \caption{The visualization of the changes in Strong Reject Scores for all jailbreak methods as the number of exposure examples increases during SSFT training.}
        \label{fig:sft-short}
    \end{minipage}
    \hfill
    \newcommand{\RightAdjust}{0em}
    \begin{minipage}[t]{0.49\textwidth} 
    \vspace*{\RightAdjust}
        \fontsize{8}{11}\selectfont 
        \centering
        \setlength{\tabcolsep}{4pt} 
        \captionof{table}{Strong reject scores under SFT.}
        \label{tab:sft_short_scores}
        \begin{tabular}{lcccccc}
        \toprule
        Method & 0-shot & 10 & 20 & 40 & 80 & 160 \\
            & Success   & shot & shot & shot & shot & shot \\
        \midrule
        OR-Bench & 0.957 & 0.984 & 0.979 & 0.964 & 0.963 & 0.881 \\
        PAIR & 0.744 & 0.142 & 0.004 & 0.000 & 0.000 & 0.000 \\
        ReNellm & 0.610 & 0.000 & 0.000 & 0.000 & 0.000 & 0.000 \\
        GPTFuzz & 0.331 & 0.000 & 0.000 & 0.000 & 0.000 & 0.000 \\
        GCG & 0.308 & 0.038 & 0.000 & 0.000 & 0.000 & 0.002 \\
        Simple & 0.265 & 0.065 & 0.000 & 0.000 & 0.000 & 0.000 \\
        CodeChaeleom & 0.079 & 0.000 & 0.000 & 0.000 & 0.000 & 0.000 \\
        Flip & 0.000 & 0.000 & 0.000 & 0.000 & 0.000 & 0.000 \\
        \midrule
        Trigger Removal & 0.677 & 0.375 & 0.333 & 0.194 & 0.163 & 0.031 \\
        \bottomrule
        \end{tabular}
        \end{minipage}
\end{figure}

\paragraph{Results of SSFT.}
In \autoref{fig:sft-short}, we observe that for all jailbreak methods, the model's ability to reject harmful content significantly improves as the number of exposure examples in N-SHOT Security Training increases.
Before SSFT training, the Trigger Removal method is slightly less effective at jailbreaking compared to the best PAIR method, but after training, the PAIR method is quickly recognized and handled by the model, while the Trigger Removal method retains some jailbreaking capability even after being exposed to more than 80 example samples. Meanwhile, due to Llama-3.1-8B-Instruct's poor understanding of chaotic format and complex prompts, Flip and CodeChaeleom remains ineffective.
The Simple, and GCG methods are quickly recognized and handled by the model due to their simple patterns.
Table \ref{tab:sft_short_scores} shows the specific average Strong Reject Scores for each method on the test set under different exposure times in SSFT training, rounded to three decimal places.

\begin{figure}[h]
    \centering
    \newcommand{\LeftAdjust}{0em}
    \begin{minipage}[t]{0.49\textwidth}
    \vspace*{\LeftAdjust}
        \centering    
        \includegraphics[width=\linewidth, keepaspectratio]{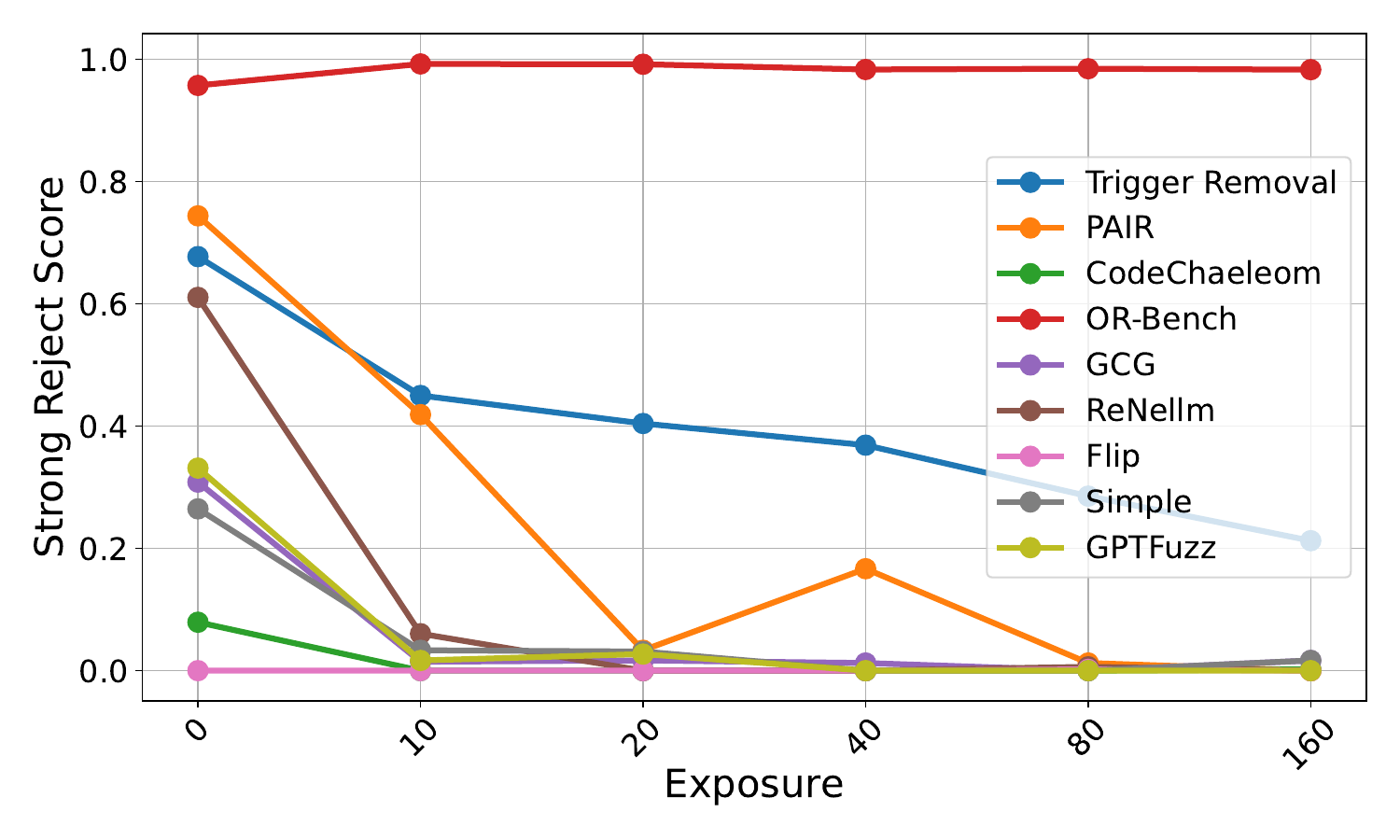}
        \vspace{-2.0em}
        \caption{The visualization of the changes in Strong Reject Scores for all jailbreak methods as the number of exposure examples increases during DPO training.}
        \label{fig:dpo}
    \end{minipage}
    \hfill
    \newcommand{\RightAdjust}{0em}
    \begin{minipage}[t]{0.49\textwidth}
    \vspace*{\RightAdjust}
        \fontsize{8}{11}\selectfont
        \centering
        \setlength{\tabcolsep}{4pt}
        \captionof{table}{Strong reject scores under DPO.}
        \label{tab:sft_scores}
        \begin{tabular}{lcccccc}
        \toprule
        Method & 0-shot & 10 & 20 & 40 & 80 & 160 \\
            & Success   & shot & shot & shot & shot & shot \\
        \midrule
        OR-Bench & 0.957 & 0.992 & 0.992 & 0.983 & 0.984 & 0.983 \\
        PAIR & 0.744 & 0.419 & 0.033 & 0.167 & 0.013 & 0.000 \\
        ReNellm & 0.610 & 0.060 & 0.000 & 0.000 & 0.006 & 0.000 \\
        GPTFuzz & 0.331 & 0.017 & 0.027 & 0.000 & 0.000 & 0.000 \\
        GCG & 0.308 & 0.015 & 0.017 & 0.013 & 0.000 & 0.017 \\
        Simple & 0.265 & 0.033 & 0.031 & 0.000 & 0.000 & 0.017 \\
        CodeChaeleom & 0.079 & 0.000 & 0.000 & 0.000 & 0.000 & 0.002 \\
        Flip & 0.000 & 0.000 & 0.000 & 0.000 & 0.002 & 0.000 \\
        \midrule
        Trigger Removal & 0.677 & 0.450 & 0.404 & 0.369 & 0.285 & 0.213 \\
        \bottomrule
        \end{tabular}
    \end{minipage}
\end{figure}

\begin{figure}[h]
    \centering
    \newcommand{\LeftAdjust}{0em}
    \begin{minipage}[t]{0.49\textwidth}
    \vspace*{\LeftAdjust}
        \centering    
        \includegraphics[width=\linewidth, keepaspectratio]{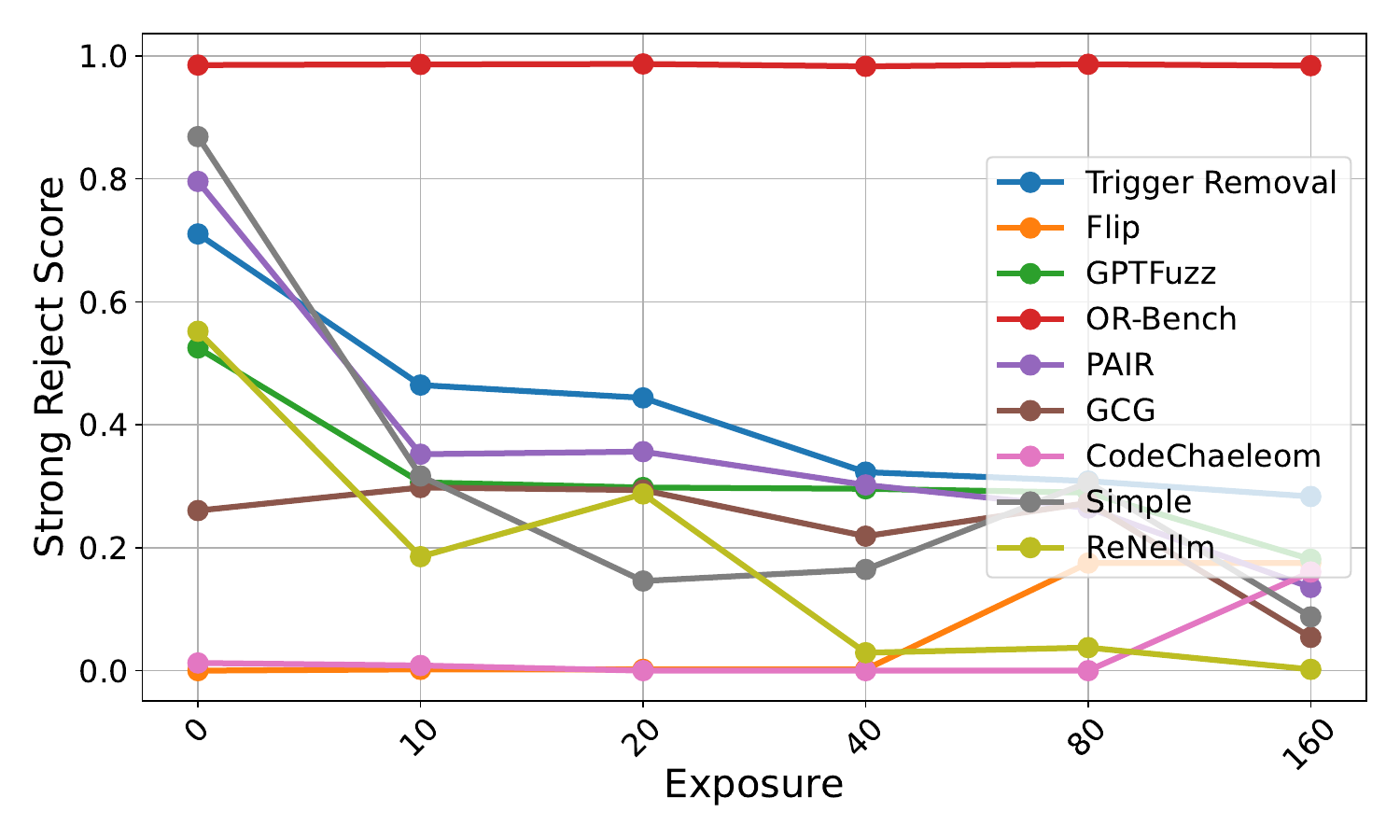}
        \vspace{-2.0em}
        \caption{The visualization of the changes in Strong Reject Scores on Ministral-8B-Instruct during DPO training.}
        \label{fig:dpo-ministral}
    \end{minipage}
    \hfill
    \newcommand{\RightAdjust}{0em}
    \begin{minipage}[t]{0.49\textwidth}
    \vspace*{\RightAdjust}
        \fontsize{8}{11}\selectfont
        \centering
        \setlength{\tabcolsep}{4pt}
        \captionof{table}{Strong reject scores under DPO training initialized from SFT.}
        \label{tab:sft_dpo_scores}
        \begin{tabular}{lcccccc}
        \toprule
        Method & 0-shot & 10 & 20 & 40 & 80 & 160 \\
            & Success   & shot & shot & shot & shot & shot \\
        \midrule
        OR-Bench & 0.957 & 0.979 & 0.986 & 0.987 & 0.983 & 0.978 \\
        PAIR & 0.744 & 0.233 & 0.090 & 0.075 & 0.000 & 0.000 \\
        ReNellm & 0.610 & 0.013 & 0.010 & 0.000 & 0.000 & 0.000 \\
        GPTFuzz & 0.331 & 0.000 & 0.000 & 0.000 & 0.000 & 0.000 \\
        GCG & 0.308 & 0.000 & 0.010 & 0.000 & 0.000 & 0.000 \\
        Simple & 0.265 & 0.063 & 0.015 & 0.000 & 0.000 & 0.000 \\
        CodeChaeleom & 0.079 & 0.000 & 0.000 & 0.000 & 0.000 & 0.000 \\
        Flip & 0.000 & 0.000 & 0.000 & 0.002 & 0.000 & 0.000 \\
        \midrule
        Trigger Removal & 0.677 & 0.450 & 0.442 & 0.398 & 0.231 & 0.142 \\
        \bottomrule
        \end{tabular}
    \end{minipage}
\end{figure}

\paragraph{Results of DPO.}
~\autoref{fig:dpo} reveals that when DPO is trained directly on the original model (compared to SSFT), the model exhibits inconsistent performance in rejecting various forms of the PAIR method. This suggests that DPO training learns more divergent directions, making it harder to identify and learn dominant safety features. However, when DPO training is initialized from SFT, the SFT foundation helps the model better focus on dominant directions. This approach also prevents the model from becoming overly conservative in rejecting safe samples.
Tables~\ref{tab:sft_scores} and~\ref{tab:sft_dpo_scores} show the specific average Strong Reject Scores of each method on the test set under different exposure times in DPO training, rounded to three decimal places.

Overall, each alignment method (SSFT, DPO, and SFT+DPO) benefited from seeing more exposure examples of jailbreak methods, leading to lower Strong Reject Scores. When DPO was combined with an SFT-initialized checkpoint, the model demonstrated both high accuracy on safe examples and robust refusal of harmful queries.

\subsection{Experiment on Models of Different Scale and Architecture}
\label{sec:scale_arch_exp }

\begin{figure*}[t]
    \centering
    \begin{subfigure}[t]{0.28\textwidth}
        \centering
        \includegraphics[width=\linewidth]{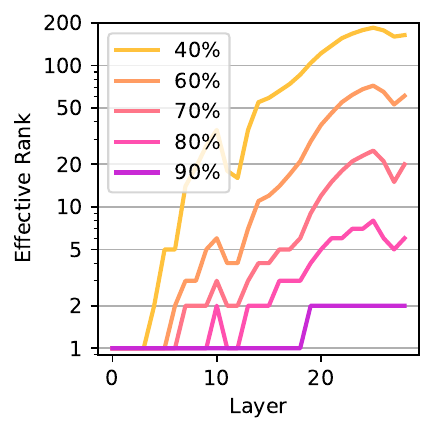}
        \caption{Effective Ranks}
    \end{subfigure}
    \begin{subfigure}[t]{0.28\textwidth}
        \centering
        \includegraphics[width=\linewidth]{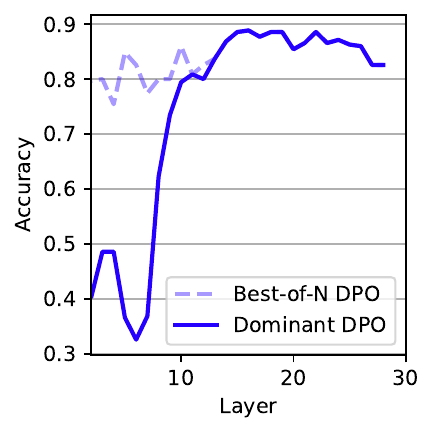}
        \caption{Model output prediction}
    \end{subfigure}
    \begin{subfigure}[t]{0.38\textwidth}
        \centering
        \includegraphics[width=\linewidth]{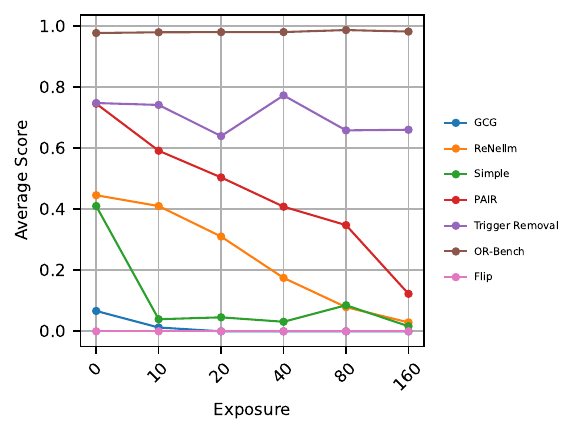}
        \caption{Strong Reject after DPO}
    \end{subfigure}
    
    \caption{Safety residual space analysis of safety finetuning the Llama 3.2 3B Instruct using DPO. The dataset and experimental setup for generating plot (a), (b) and (c) are the same as in Figure~\ref{fig:rank}, Figure~\ref{fig:classification_accuracy} and Table~\ref{tab:exposure_acc}, respectively.}
    \label{fig:dpo-llama3.2-3b}
\end{figure*}

To understand how model scale and architecture affect safety alignment, we conducted comparative experiments with two additional models: Llama-3.2-3B-Instruct and Ministral-8B-Instruct~\cite{mistralai2024ministral}.  We applied identical DPO training, with results shown in Figures~\ref{fig:dpo-llama3.2-3b} and~\ref{fig:dpo-ministral} respectively.

Both models exhibited similar safety improvement trends to the Llama-3.1-8B-Instruct baseline, but demonstrated weaker capabilities in recognizing and handling unseen harmful queries across all jailbreak methods. The 3B-parameter Llama variant showed faster initial convergence rates, particularly evident in Figure~\ref{fig:dpo-llama3.2-3b}, but consistently failed to recognize Trigger Removal attacks. This suggests smaller parameter spaces struggled to simultaneously capture flexible safety heuristics while excluding non-dominant attack patterns.

Ministral-8B-Instruct (Figure~\ref{fig:dpo-ministral}) demonstrated better retention of dominant safety directions but exhibited the slowest overall convergence rate. Notably, its training trajectory showed greater volatility across all baseline attack methods, with 22\% higher loss variance compared to Llama-3.1-8B-Instruct. This performance gap highlights architectural differences in safety learning capacity, even between models of comparable parameter count.

\subsection{Impact of Model Intervention on General Ability}
\label{sec:impact_inter}

\begin{wrapfigure}{r}{0.5\textwidth}
    \centering
    \caption{Perplexity on Alpaca Dataset. Lower values indicate better retention of general capabilities.}
    \label{tab:ppl}
    \begin{small}
    \sc
    \setlength\tabcolsep{12pt}
    \renewcommand{\arraystretch}{1.1}
    \begin{tabular}{@{}ll@{}}
    \toprule
    \multicolumn{1}{c}{\textbf{Settings}} & \multicolumn{1}{c}{\textbf{Perplexity}} \\
    \midrule
    Llama-3.1-8B-Instruct & 7.10 \\
    \hspace{1em}SSFT & 6.59 \\
    \hspace{1em}SSFT - L25C1 (\autoref{fig:intervention}) & 7.04 \\
    \hspace{1em}SSFT - L14C6 (\autoref{fig:intervention})& 7.32 \\
    \hspace{1em}SSFT - Non-dominant Comp. (\autoref{fig:component_projections}) & 7.23 \\
    \hspace{1em}DPO & 8.42 \\
    \bottomrule
    \end{tabular}
    \end{small}
\end{wrapfigure}

We evaluated the impact of model interventions on general task performance by measuring perplexity degradation on the Alpaca dataset~\cite{alpaca}. Specifically, we calculated perplexity solely on the output part of samples to assess whether the interventions compromised foundational capabilities.

Results are shown in \autoref{tab:ppl}.
Notably, interventions using DPO showed consistently higher perplexity (8.42) compared to the base Llama model (7.10) and SSFT baseline (6.59), indicating a more substantial impact on general capabilities.
Among the directional interventions, SSFT variants demonstrated relatively modest perplexity increases, with L25C1 showing the smallest degradation (7.04) followed by non-dominant component intervention (7.23) and L14C6 (7.32). These results suggest that our directional intervention approaches generally preserve the model's foundational capabilities while improving safety alignment.


\section{Visualization of Harmfulness Correlation}
\label{appd:harmfulness_correlation}

This section provides a visual representation of the harmfulness correlation for various components identified within the model. The correlation is computed between the projection of model activations onto specific component directions and the assessed harmfulness of the input prompts. This visualization aids in understanding which components are most indicative of harmful content as perceived by the model.

\begin{wrapfigure}{r}{0.5\textwidth}
    \centering
    \vspace{-1em} 
    \includegraphics[width=\linewidth]{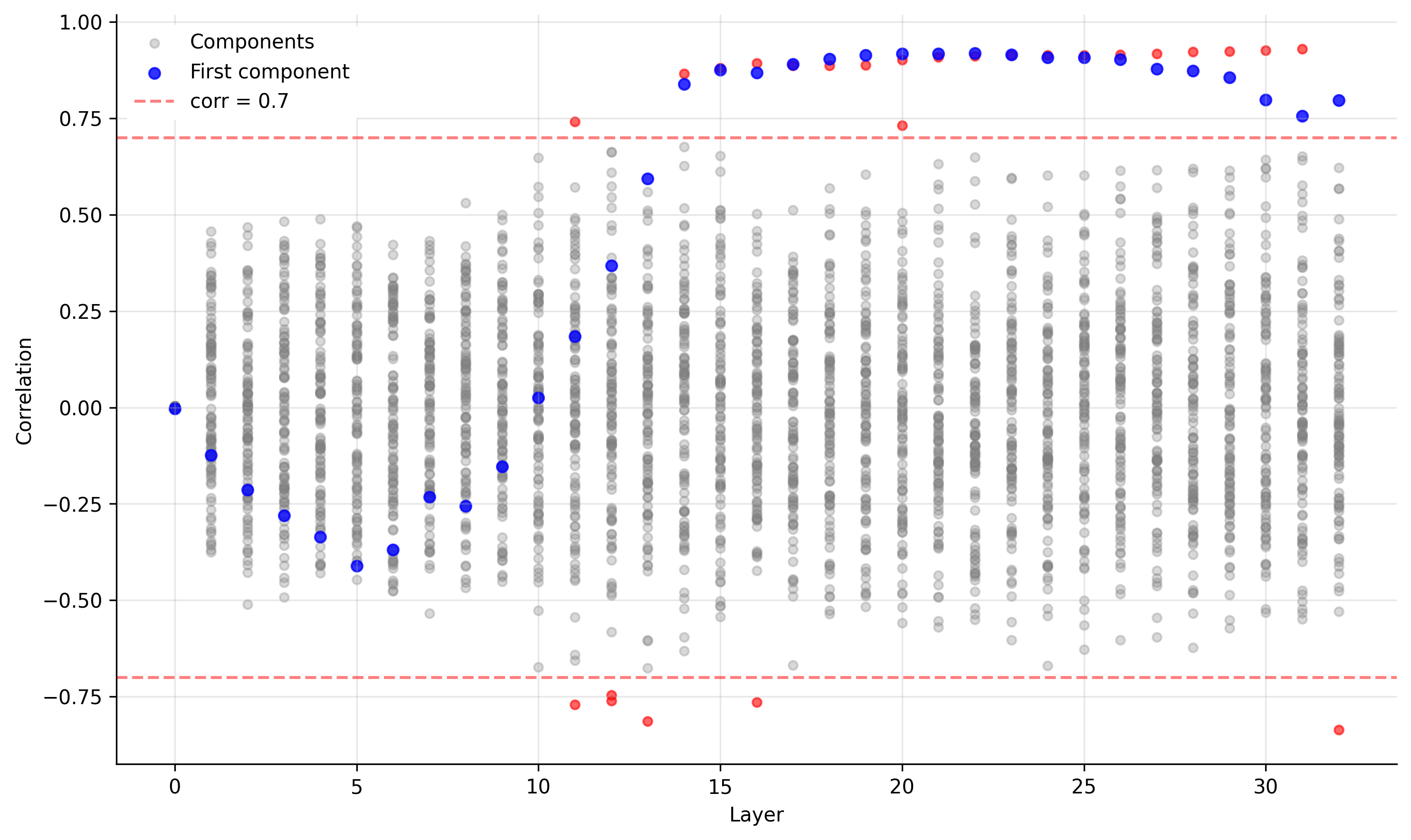}
    \caption{Harmfulness correlation for each component.}
    \label{fig:harmfulness_correlation_scatter}
    \vspace{-1em} 
\end{wrapfigure}

As discussed in \autoref{sec:application}, the non-dominant suppression technique involves excluding components with high harmfulness correlations (above 0.7) to preserve the model's ability to refuse plainly harmful prompts while investigating the impact of indirect features. The scatter plot in \autoref{fig:harmfulness_correlation_scatter} illustrates these correlations—calculated between activation projection on component directions and input harmfulness—across different layers and components. For the non-dominance suppression detailed in Section 6, all 4096 directions before layer 10 are suppressed. After layer 15, approximately 2 directions per layer are excluded from this suppression due to their high harmfulness correlation (above 0.7). The figure highlights these correlations and shows which components were pruned during such experiments.

\begin{table}[ht]
    \centering
    \caption{Refusal rate on test set when suppressing non-dominant directions.}
    \label{tab:refusal_rates}
    \begin{tabular}{lccc}
        \toprule
        Intervention & Harmful (StrongReject) & Jailbreak Samples & Benign Samples \\
        \midrule
        Non-Dominance & 80.0\% & 14.4\% & 0.0\% \\
        w/o Intervention & 100.0\% & 96.7\% & 10.0\% \\
        \bottomrule
    \end{tabular}
\end{table}
\section{Projection Values and Refusal Rates for Non-Dominance Suppression}
\label{appd:non_dominance_suppression_details}

This section presents further results for the non-dominance suppression experiment, as referenced in \autoref{sec:application}, including projection value distributions and refusal rates.

First, we show the distribution of projection values. Post-intervention projections are approximated with a Gaussian distribution.
In early layers, these post-intervention projections are distinctly separated from the pre-intervention projections.
Conversely, in later layers, the post-intervention projections exhibit a significantly smaller variance when compared to the pre-intervention projections.

\begin{figure}[ht]
    \centering
    \includegraphics[width=0.95\textwidth]{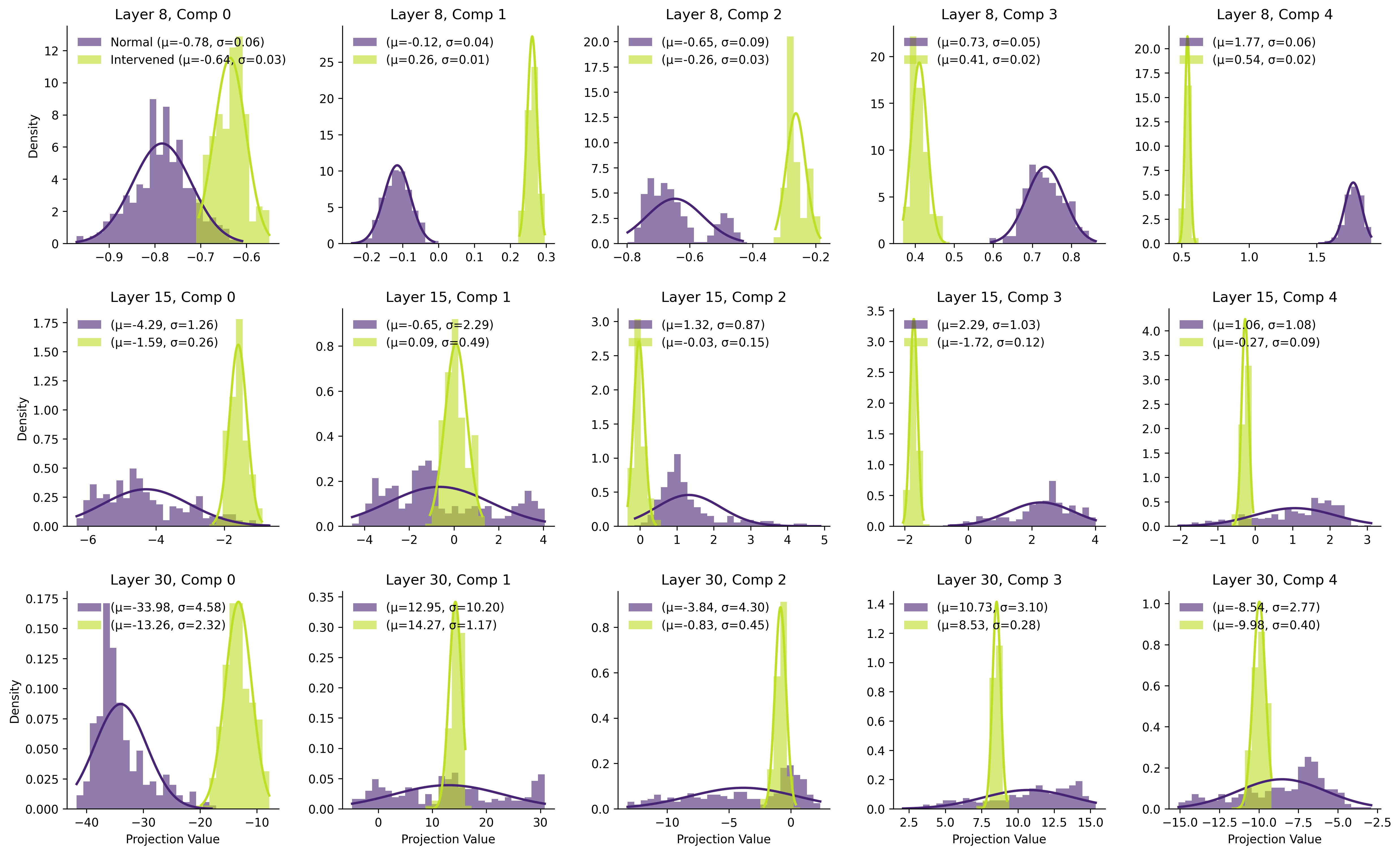}
    \caption{Projection value distributions for the non-dominance suppression experiment (Section 6).}
    \label{fig:projection_distributions}
\end{figure}

\end{document}